\documentclass[journal]{IEEEtran}%duoble
\newcommand{\ignore}[1]{}
\usepackage[utf8]{inputenc}
\usepackage[T1]{fontenc}
\usepackage{graphicx}
\usepackage{amsmath}
\usepackage{amssymb}
\usepackage{booktabs}
\usepackage{times}
\usepackage{epsfig}
\usepackage{cite}

\usepackage{colortbl}
\usepackage{algorithm}  
\usepackage{algpseudocode}
\usepackage{xcolor}
\usepackage{pifont}
\usepackage{multirow}
\usepackage{bbm}
\usepackage{mathtools}
\newcommand{\cmark}{\ding{51}}
\newcommand{\xmark}{\ding{55}}
\newcommand{\etal}{\textit{et al}. }

\begin{document}
%
% paper title
% can use linebreaks \\ within to get better formatting as desired
\title{Plug-and-Play Regulators for Image-Text Matching}
\author{Haiwen~Diao,
        Ying~Zhang,
        Wei~Liu,
        Xiang Ruan
        and~Huchuan~Lu 
\thanks{
% Copyright (c) 2022 IEEE. Personal use of this material is permitted. However, permission to use this material for any other purposes must  be obtained from the IEEE by sending an email to \underline{pubs-permissions@ieee.org}.
This work was supported in part by the National Key R\&D Program of China under Grant No.2018AAA0102001 and National Natural Science Foundation of China under grant No.62293542, U1903215 and the Fundamental Research Funds for the Central Universities No.DUT22ZD210.

H. Diao and H. Lu are with School of Information and Communication Engineering, Dalian University of Technology, Dalian, 116024, China.  (Email: diaohw@mail.dlut.edu.cn; lhchuan@dlut.edu.cn).
Y. Zhang and W. Liu are with Tencent Holdings Limited, Shenzhen, 518054, China.
(Email: yinggzhang@tencent.com; wl2223@columbia.edu).
X. Ruan is with Tiwaki Company Limited, Kusatsu, 5258577, Japan.
(Email: ruanxiang@tiwaki.com).
}}
% make the title area
\maketitle
% The paper headers
\markboth{IEEE Transactions on Image Processing}{}

\begin{abstract}
Exploiting fine-grained correspondence and visual-semantic alignments has shown great potential in image-text matching. Generally, recent approaches first employ a cross-modal attention unit to capture latent region-word interactions, and then integrate all the alignments to obtain the final similarity. However, most of them adopt one-time forward association or aggregation strategies with complex architectures or additional information, while ignoring the regulation ability of network feedback. In this paper, we develop two simple but quite effective regulators which efficiently encode the message output to automatically contextualize and aggregate cross-modal representations. Specifically, we propose (i) a Recurrent Correspondence Regulator (RCR) which facilitates the cross-modal attention unit progressively with adaptive attention factors to capture more flexible correspondence, and (ii) a Recurrent Aggregation Regulator (RAR) which adjusts the aggregation weights repeatedly to increasingly emphasize important alignments and dilute unimportant ones. Besides, it is interesting that RCR and RAR are “plug-and-play”: both of them can be incorporated into many frameworks based on cross-modal interaction to obtain significant benefits, and their cooperation achieves further improvements. Extensive experiments on MSCOCO and Flickr30K datasets validate that they can bring an impressive and consistent R@1 gain on multiple models, confirming the general effectiveness and generalization ability of the proposed methods.
\end{abstract}
\IEEEpeerreviewmaketitle

%--------------------------------------------------------------------------
%--------------------------------------------------------------------------

\section{Introduction}
Exploiting the interactions between vision and language has attracted great interests in past decades, and various applications have sprouted to associate vision and text such as video-text retrieval~\cite{HGR,EHLS,HIT}, visual question answering~\cite{DAN}, image captioning~\cite{BU_TDA}, visual grounding~\cite{LGGAN}, and visual commonsense reasoning~\cite{VCR}. Among them, image-text matching involves the transmission and measurement of the cross-modal information, and provides great help for other tasks, making it become an important branch in the computer vision research area.

\begin{figure}[htpb]
	\centering
	\begin{tabular}{@{}c}
    	\includegraphics[width=0.98\linewidth, height=0.4\linewidth,trim=0 310 390 0,clip]{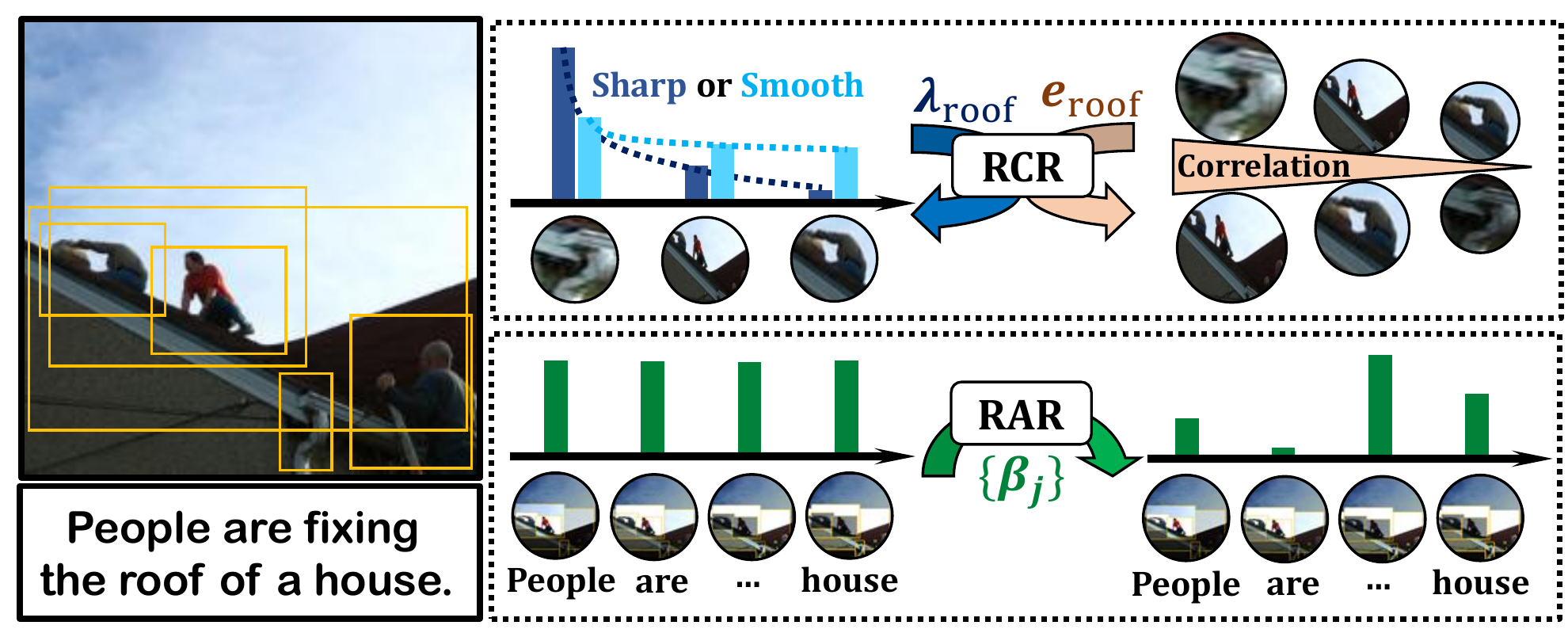} 
	\end{tabular}
	\caption{Illustration of the proposed regulators. RCR progressively produces a plausible attention distribution between the word \texttt{"roof"} and its corresponding regions by adjusting the temperature $\lambda$ and channel-wise factor $\boldsymbol{e}$, while RAR constantly highlights significant alignments attended by each word and boosts itself step by step for more comprehensive aggregation.}
	\label{fig:motivation}
\end{figure}

Great efforts have been made to accurately establish the relationship between visual and textual observations. Early works such as~\cite{DSPE,VSE++,DAN,SCO,GXN,PVSE,SAEM,VSRN,SGM,CVSE,GPO} attempted to map the whole image and the full sentence into a joint embedding space, where the similarity between different modalities can be directly measured. These approaches capture the global correspondence between an image and a sentence, while ignoring the importance of exploring fine-grained interactions across two modalities. 
To solve this problem, Lee \etal~\cite{SCAN} proposed a cross-modal attention mechanism to explore region-word correspondences, which achieved impressive bi-directional retrieval results. Following it, many researchers are devoted to exploiting more accurate latent correspondence by either improving the cross-attention unit~\cite{BFAN,MMCA,RDAN}, or enhancing the cross-modal embeddings~\cite{PFAN,DP-RNN,ACME,CAAN,IMRAM}. For example, Hu \etal~\cite{RDAN} utilized a visual-semantic relation CNN to refine region-word interactions, while Chen \etal~\cite{DP-RNN} reinforced the semantically related objects via encoding the image regions with a Recurrent Neural Network in order of the matching-word positions.
Another thread of works focus more on the matching stage and infer the final similarity via aggregating all the alignments. Most existing approaches adopted the strategy of averaging all the cosine similarities between local alignments~\cite{SCAN, PFAN, BFAN, RDAN, IMRAM}, which has achieved satisfying performance but remains far from being optimal. Chen \etal~\cite{DP-RNN} introduced the self-attention mechanism to weight the image-word and object-text similarities separately, while \cite{GSMN,SGRAF} performed graph reasoning on the matching vectors to gain more benefits over the simple cosine aggregation.

However, we observe that most approaches targeting cross-modal interactions focus on developing the interactive capability with one-time forward architectures~\cite{RDAN,BFAN} or incorporating various information, such as intra-inter relationship~\cite{sm-LSTM,CAMP,MMCA,CAAN}, object position~\cite{PFAN,GSMN}, semantic order~\cite{DP-RNN}, phrase structure~\cite{R-SCAN,GSMN}, while neglecting the regulation ability of network itself which learns from the message feedback and then leads to accurate and dynamic optimization. On the other hand, most of the above methods adopt the equivalent cross-modal alignment aggregation strategy for diverse region/word semantics and positive/negative pairs, lacking the ability to refine undesirable associations and capture complicated matching patterns. In this paper, we introduce a regulator mechanism defined by ~\cite{Black-Box,RegNet,AdaFR,Feedpixel} where the network can be improved by adaptively optimizing the forward learning process with plausible backward feedback loops, and validate that an elaborate regulation operation can make a vast difference in obtaining accurate interactions and conducting optimal aggregations across modalities requiring no additional data and complicated structures.

To be more specific, we propose a Recurrent Correspondence Regulator (RCR) and a Recurrent Aggregation Regulator (RAR) to progressively promote the image-text matching process, as shown in Fig.~\ref{fig:motivation}. The RCR learns adaptive attention factors for each specific word/region to refine the cross-modal attention unit iteratively, acquiring more plausible attention distributions for semantically diverse words/regions in various image-text pairs. The RAR starts with averaging all the alignments and then updates the aggregation weights guided by the aggregated alignment in the previous step, which increasingly emphasizes important alignments and gradually reduces the interference of unimportant ones to predict more precise similarity scores. An important and attractive property of the proposed RCR and RAR is “plug-and-play”: both of them can be seamlessly inserted into many existing methods based on cross-modal interaction to achieve remarkable improvements, and their cooperation brings greater benefits. Moreover, we experimentally verify that even with the simplest framework, the plug-ins of RCR and RAR enable the model~\cite{SCAN} to achieve state-of-the-art results on MSCOCO and Flickr30K.
In summary, our main contributions are three-fold:
\begin{itemize}
	\item We propose a Recurrent Correspondence Regulator (RCR) to dynamically renew the cross-attention unit for better correspondence exploitation. It learns adaptive attention factors for each word/region to generate a more plausible attention distribution in accordance with its semantics and associated image-text pairs.
	\item We propose a Recurrent Aggregation Regulator (RAR) to repeatedly calibrate the weights for more discriminative similarity aggregation. It progressively reweighs word/region-attended alignments directed by earlier guidance alignment to highlight more important alignments.
	\item The RCR and RAR can be applied to various approaches for image-text matching separately or jointly to achieve significant improvements, indicating the effectiveness and generalization ability of the proposed approach.
\end{itemize}

%------------------------------------------------------------
\section{Related work}
\subsection{Cross-modal Attention}
Cross-modal attention is first developed by Lee \etal~\cite{SCAN} to discover all possible word-region alignments for image-text matching.
With spectacular achievements, it attracts numerous researchers to make further explorations on enhancing cross-modal embeddings or improving attention units. 
Specifically, the former works attempt to facilitate word-region correspondence by enriching the instance features with region position~\cite{PFAN}, semantic order~\cite{DP-RNN}, scene graph~\cite{HOAD,HAT} and intra-inter correlation~\cite{MMCA}, while the latter methods directly develop more fine-grained interactions across modalities, such as relation CNN~\cite{RDAN}, focal attention~\cite{BFAN} and cross-graph attention~\cite{HAT,GSMN}.
In particular, some works~\cite{HAT,HOAD,GSMN} introduced scene graphs with explicit attribute and relation information, and then constructed an inter-graph attention between graph nodes. Besides, Qu \etal~\cite{DIME} developed a routing mechanism to realize dynamic modality interaction, while Zhang \etal~\cite{NAAF} exploited both the matched and mismatched effects for a comprehensive image-text relationship.
Compared with previous works recurrently updating query features~\cite{IMRAM}, instance fusion~\cite{sm-LSTM,DAN}, and context enhancement~\cite{MMCA}, the RCR directly adjusts the attention factors including the channel-wise weight vector and the softmax temperature, enabling attention weights to adapt to diverse semantic regions/words with different word/region sets from various image-text pairs. To be specific, for positive image-text pairs, the attention weights between each word/region and its corresponding regions/words are more precise and discriminative by the RCR, leading to a tighter distance between image and text. More importantly for negative pairs with completely irrelevant instances, the inner-product-like weights between the paired features adopted by the existing attention designs still emphasize the closest contents across modalities and capture the so-called "region-word correlations" in the latent space, inevitably increasing the image-text similarity and reducing the gap from the positive pairs. In contrast, the RCR progressively adjusts word-region relevance by the learned attention factors to generate appropriate attention distributions targeting diverse regions/words, meanwhile decoupling the attention weights from the final similarity measurement and producing larger gaps between matched and unmatched pairs.

\subsection{Similarity Aggregation}
Existing approaches~\cite{VSRN,SGM,WCGL,AME,GPO,CODER} based on mono-modal representation map the image and text features into a joint space and adopt the cosine distance as the measurement, while a great many methods~\cite{SCAN, PFAN, BFAN, RDAN, IMRAM} based on cross-modal interaction first obtain the pairwise features across modalities and then employ the average operation to fuse the cosine similarity between all the word-region alignments. 
Considering that various instances and hierarchical relevance have different importance in characterizing the cross-modality relations, Chen \etal~\cite{DP-RNN} designed a self-attention module to integrate all cosine distances attended by regions or words, while Ji \etal~\cite{SHAN} explored both fragment-wise and context-wise similarity scores to yield sufficient visual-semantic alignments between image and text.
Besides for more powerful distance representations, some methods~\cite{GSMN,SGRAF,CMCAN} introduced a vector-based similarity function and performed the matching pattern with graph reasoning, which have achieved great improvements at the cost of high complexity. 
Note that Liu \etal~\cite{GSMN} not only needs to parse additional visual/textual graphs, but also fails to measure diverse alignments by aggregating them with average pooling operation. 
Moreover, Diao \etal~\cite{SGRAF} requires a high-quality holistic alignment to better guide the integration procedure of fragment-wise alignments. 
In contrast, our RAR employs a recurrent aggregation process without any extra supplement and precondition, and validates that an iterative guidance alignment encoding early matching information can yields more appropriate weights and effectively facilitate the aggregation process for various alignments.

\subsection{Plug-and-Play Methods}
The modules that enable efficient integration into main frameworks are referred to as "plug-and-play" approaches. In recent years, the plug-and-play manners have attracted more attention in various fields, including image restoration~\cite{IDBP,H-PnP,H-PnP_extension,DPIR}, visual captioning~\cite{PickNet,RecNet}, visual question answering~\cite{RAGAN,CSS}, and video-text matching~\cite{MAC,RegionLearner}. By decoupling a specific problem from overall optimization objectives, they greatly simplify the integration process of each module, and improve flexibility and generalizability on new frameworks, thus accelerating the developments over other more sophisticated applications. It is worth noting that the method most relevant to us is GPO~\cite{GPO}, which attempts to improve the mono-modal feature encoders and learn the best pooling strategy to integrate mono-modal instance features into a holistic embedding, while our regulators aim to generalize over various cross-modal interaction methods and promote multi-modal attention and similarity aggregation.

\section{Background}
In this section, we briefly review the Stacked Cross Attention Network (SCAN)~\cite{SCAN}, which serves as the pioneer in exploiting word-region correspondences and alignments for image-text matching task. The whole architecture consists of four aspects: Feature Extraction, Cross-modal Attention, Similarity Computation, and Objective Function.

\subsection{Feature Extraction}
\label{secFE}
\textbf{Image Representation.} 
Given an image, the Faster R-CNN~\cite{FasterR-CNN} model pretrained on Visual Genome~\cite{VisualGenome} is first utilized to detect $K$ salient regions with bottom-up attention~\cite{BU_TDA}, followed with a linear layer transforming each region feature into a $d$-dimensional vector. Therefore the image is encoded as a set of region features $\boldsymbol{V} = \{\boldsymbol{v}_{1},...,\boldsymbol{v}_{K}\}$, with $\boldsymbol{v}_{i}\in \mathbb{R}^{d}$ denoting the feature of $i$-th region.

\textbf{Text Representation.}
Given a sentence with $L$ words, we represent each word with a one-hot vector by random initialization, and map it into a 300-dimensional word embedding, followed by a bi-directional GRU~\cite{Bi-GRU} to integrate the bidirectional contextual information. The final text feature is computed by averaging the forward and backward hidden states to obtain $\boldsymbol{T} = \{\boldsymbol{t}_{1},...,\boldsymbol{t}_{L}\}$, and $\boldsymbol{t}_{j}\in \mathbb{R}^{d}$ indicates the representation of $j$-th word.

\subsection{Cross-modal Attention}
Here, we only depict the text-to-image (T2I) attention in detail, and the image-to-text attention (I2T) performs similar operations. Given a set of region features $\boldsymbol{V} = \{\boldsymbol{v}_{1},...,\boldsymbol{v}_{K}\}$ and word features $\boldsymbol{T} = \{\boldsymbol{t}_{1},...,\boldsymbol{t}_{L}\}$, the attention unit first computes the cosine similarities between all word-region pairs:
\begin{equation}
\label{eq:cij}
    c_{ij}^{(0)} = R(\boldsymbol{v}_{i}, \boldsymbol{t}_{j}|\mathbbm{1}^{d}) = \frac{\boldsymbol{v}_{i}^{\top}(\mathbbm{1}^{d} \odot \boldsymbol{t}_{j})}{\|\boldsymbol{v}_{i}\|\|\boldsymbol{t}_{j}\|} \ ,
\end{equation}
where $R(\cdot,\cdot|\mathbbm{1}^{d})$ indicates the cosine similarity function which computes the inner product weighted by the channel-wise all-ones vector $\mathbbm{1}^{d}$. The attention weights are then calculated by a softmax function as
\begin{equation}
\label{eq:alphaij}
    \alpha_{ij}^{(0)} = \frac{exp({\lambda}\bar{c}_{ij}^{(0)})}{{\sum}_{i=1}^{K}exp({\lambda}\bar{c}_{ij}^{(0)})} \ ,
\end{equation} 
where $\bar{c}_{ij}^{(0)}$ = ${{\left [ c_{ij}^{(0)} \right ]}_{+}}/{\sqrt{{\sum}_{j=1}^{L}{\left [ c_{ij}^{(0)} \right ]}_{+}^{2}}}$ with $\left [ x \right ]_{+}$ = $max(x, 0)$, and $\lambda$ is the temperature of the softmax function. 
Here, ${\alpha}_{ij}^{(0)}$ is the normalized attention weight capturing the correspondence between the $j$-th word and its related regions, and thus the image feature attended by each word can be obtained via
\begin{equation}
\label{eq:hatvj}
    \hat{\boldsymbol{v}}_{j}^{(0)} = \sum \limits_{i=1}^{K}{\alpha}_{ij}^{(0)}\boldsymbol{v}_{i} \ .
\end{equation}

For simplicity, we define the cross-modal attention unit as
\begin{equation}
\label{eq:cma}
    \hat{\boldsymbol{v}}_j^{(0)} = \mathbf{CMA}(\boldsymbol{t}_j,
    \boldsymbol{V}|\mathbbm{1}^{d},\lambda) \ ,
\end{equation}
where the integrated image feature $\hat{\boldsymbol{v}}_{j}^{(0)}$ represents the related image regions with respect to $j$-th word under the fixed attention factors, including a channel-wise weight vector $\mathbbm{1}^{d}$ and a softmax temperature $\lambda$.

\subsection{Similarity Computation}
The final image-text similarity is computed by averaging all the cosine similarities between $\hat{\boldsymbol{v}}_{j}^{(0)}$ and $\boldsymbol{t}_{j}$ as
\begin{equation}
\label{eq:st2i}
    \mathcal{S}_{T2I} = \frac{1}{L}\sum_{j=1}^{L} R(\hat{\boldsymbol{v}}_{j}^{(0)}, \boldsymbol{t}_{j} |\mathbbm{1}^{d}) \ .
\end{equation} 
Similarly, the predicted similarity score by I2T attention is denoted as $\mathcal{S}_{I2T}$, and the combination of these two scores usually produces greater retrieval results. 

\subsection{Objective Function}
Given a matched image-text pair ($\boldsymbol{V},\boldsymbol{T}$), the hard ranking loss~\cite{VSE++} with online negative mining only takes account of the nearest negatives ($\tilde{\boldsymbol{T}}, \tilde{\boldsymbol{V}}$) within a mini-batch $\mathcal{D}$. The similarity of positive pairs should be higher than that of negative pairs by a fixed margin value $\gamma$, which is formulated as
\begin{equation}
\label{eq:losfun}
\begin{split}
\mathcal{L} = \sum_{(\boldsymbol{V}, \boldsymbol{T})\in\mathcal{D}}
&[\gamma + \mathcal{S}(\boldsymbol{V}, \tilde{\boldsymbol{T}}) - \mathcal{S}(\boldsymbol{V}, \boldsymbol{T})]_{+} \\
+\ &[\gamma + \mathcal{S}(\tilde{\boldsymbol{V}}, \boldsymbol{T}) - \mathcal{S}(\boldsymbol{V}, \boldsymbol{T})]_{+} \ ,
\end{split}
\end{equation}
where $\mathcal{S}(\cdot,\cdot)$ represents the matching score of an image-text pair computed by the aforementioned network.

\section{Methodology}
In this section, we will elaborate on the proposed Recurrent Correspondence Regulator (RCR) and Recurrent Aggregation Regulator (RAR) based on the cross-modal attention unit from SCAN~\cite{SCAN}. 
These two regulators can effectively explore the regulatory capacity of the network itself and in turn significantly facilitate the learning process by exploiting the well-designed alignment feedback. 
For simplicity, we take the T2I attention to describe the proposed regulation strategies, which can be applied to the I2T attention in the same way.

\begin{figure}[t!]
	\centering
	\begin{tabular}{@{}c}
		\includegraphics[width=0.98\linewidth, height=0.34\linewidth,trim= 0 330 300 0,clip]{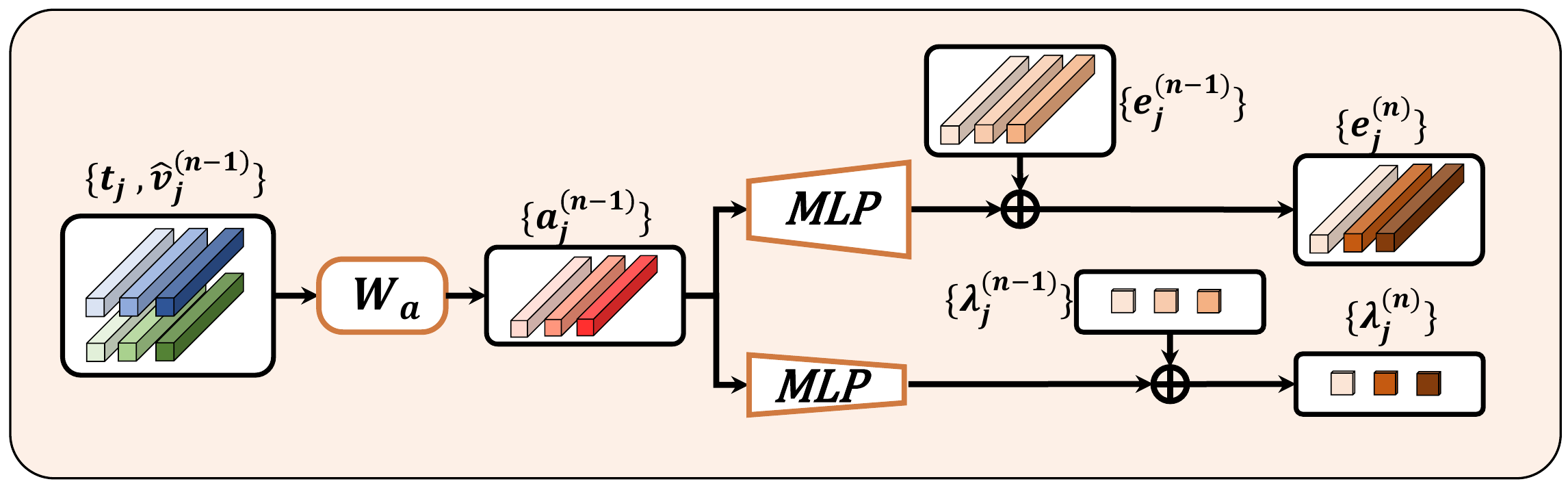}
	\end{tabular}
	\caption{Illustration of RCR that refines the cross-modal interactions via learning new channel-wise weight vectors and softmax temperature.}
	\label{fig:RCR}
\end{figure}

\subsection{Recurrent Correspondence Regulator} 
The Recurrent Correspondence Regulator (RCR) learns specific attention factors for each word in a recurrent manner, aiming to refine the correspondence between each word and all the regions. In Fig.~\ref{fig:RCR} with the word feature $\boldsymbol{t}_{j}$ and its related image feature $\hat{\boldsymbol{v}}_{j}^{(0)}$, we first construct the alignment vector $\boldsymbol{a}_{j}^{(0)}$ following~\cite{GRN19} with respect to $\boldsymbol{t}_{j}$ via
\begin{equation}
\label{eq:aj}
\boldsymbol{a}_{j}^{(0)} = \frac{\boldsymbol{W}_{a}{\left | \boldsymbol{t}_{j}- \hat{\boldsymbol{v}}_{j}^{(0)} \right |}^{2}}{{\left \| \boldsymbol{W}_{a}{\left | \boldsymbol{t}_{j}- \hat{\boldsymbol{v}}_{j}^{(0)} \right |}^{2} \right \|}_{2}} \ ,
\end{equation}
where $\boldsymbol{W}_{a}\in \mathbb{R}^{m\times d}$ is a linear transformation, and $\boldsymbol{a}_{j}^{(0)} \in \mathbb{R}^{m}$ encodes the element-wise differences and fine-grained relationships between $\boldsymbol{t}_{j}$ and $\hat{\boldsymbol{v}}_{j}^{(0)}$. With the comprehensive alignment encoding across two modalities, the alignment vector $\boldsymbol{a}_{j}^{(0)}$ is utilized to learn adaptive attention factors with multi-layer perceptron (MLP) for the next word-region interaction:
\begin{equation}
\label{eq:factors}
\begin{split}
    \boldsymbol{e}_{j}^{(1)} &= \left [ \sigma(\boldsymbol{W}_{e^{'}}(\sigma(\boldsymbol{W}_{e}\boldsymbol{a}_{j}^{(0)}+\boldsymbol{b}_{e}))+\boldsymbol{b}_{e^{'}}) + \boldsymbol{e}_{j}^{(0)} \right]_{-1}^{+1} \ , \\
    \lambda_{j}^{(1)} &= \left [ \boldsymbol{W}_{\lambda^{'}}(\sigma(\boldsymbol{W}_{\lambda}\boldsymbol{a}_{j}^{(0)}+\boldsymbol{b}_{\lambda}))+\boldsymbol{b}_{\lambda^{'}}+\lambda_{j}^{(0)} \right]_{+} \ ,
\end{split}
\end{equation}
where $\boldsymbol{W}_{\{\cdot\}}$ and $\boldsymbol{b}_{\{\cdot\}}$ are several learnable parameters, $\sigma$(·) indicates the tanh activation, and $[x]_{-1}^{+1}$ clips the value $x$ to be within $[-1, +1]$. Note that each value of the vector $\boldsymbol{e}_{j}^{(1)}$ ranges from -1 to 1, reweighing the channel-wise negative or positive correlation between $\boldsymbol{t}_{j}$ and $\hat{\boldsymbol{v}}_{j}^{(0)}$. Besides, the scalar $\lambda_{j}^{(1)}$ belongs to $[0, +\infty)$, controlling the word-wise smoothness or sharpness of attention distribution in relation to $\boldsymbol{t}_{j}$.

Then we refine the word-region correspondence in the next step by separately reformulating Eq.~\eqref{eq:cij}-\eqref{eq:hatvj} as
\begin{equation}
\label{eq:updatecij}
    c_{ij}^{(1)} = R(\boldsymbol{v}_{i}, \boldsymbol{t}_{j}|\boldsymbol{e}_{j}^{(1)}) = \frac{\boldsymbol{v}_{i}^{\top} (\boldsymbol{e}_{j}^{(1)} \odot \boldsymbol{t}_{j})}{\|\boldsymbol{v}_{i}\|\|\boldsymbol{t}_{j}\|} \ ,
\end{equation}

\begin{equation}
\label{eq:updatealphaij}
    \alpha_{ij}^{(1)} = \frac{exp({\lambda_{j}^{(1)}}\bar{c}_{ij}^{(1)})}{{\sum}_{i=1}^{K}exp({\lambda_{j}^{(1)}}\bar{c}_{ij}^{(1)})} \ ,
\end{equation}

\begin{equation}
\label{eq:updateahatvj}
    \hat{\boldsymbol{v}}_{j}^{(1)} = \sum \limits_{i=1}^{K}{\alpha}_{ij}^{(1)}\boldsymbol{v}_{i} \ ,
\end{equation}
where $\boldsymbol{e}_{j}^{(1)} \in \mathbb{R}^{d}$ is the adaptive channel-wise weight vector learned to rectify the correlation between $\boldsymbol{v}_{i}$ and $\boldsymbol{t}_{j}$, which we term as $R(\cdot, \cdot|\boldsymbol{e}_{j}^{(1)})$, and $\lambda_{j}^{(1)} \in \mathbb{R}^{1}$ is the adaptive word-wise softmax temperature which adjusts the attention distribution. $\hat{\boldsymbol{v}}_{j}^{(1)}$ is the new integrated image feature with respect to $j$-th word. $\odot$ denotes the element-wise multiplication. 

The above equations illustrate how to update the word-attended image feature via learning two new attention factors in a single step. Similarly, the regulation process can be extended to multiple runs for further refinement.
In this paper, we simplify the RCR as 
\begin{equation}
\label{eq:rcr}
    \boldsymbol{e}_j^{(n)}, \lambda_j^{(n)} = \mathbf{RCR}(\boldsymbol{t}_j,  \hat{\boldsymbol{v}}_{j}^{(n-1)}, \boldsymbol{e}_j^{(n-1)}, \lambda_j^{(n-1)}) \ ,
\end{equation}
and plug it into the cross-modal attention unit via
\begin{equation}
\label{eq:camrcr}
\resizebox{.88\hsize}{!}{$ \hat{\boldsymbol{v}}_j^{(n)} = \mathbf{CMA}(\boldsymbol{t}_j, \boldsymbol{V}| \mathbf{RCR}(\boldsymbol{t}_j, \hat{\boldsymbol{v}}_j^{(n-1)},\boldsymbol{e}_j^{(n-1)}, \lambda_j^{(n-1)}))$} ,
\end{equation}
where $\hat{\boldsymbol{v}}_j^{(n)}$ is the updated image feature attended by $j$-th word in the $n$-th regulation step.

\begin{figure}[t!]
	\centering
	\begin{tabular}{@{}c}
		\includegraphics[width=0.98\linewidth, height=0.34\linewidth,trim= 0 330 290 0,clip]{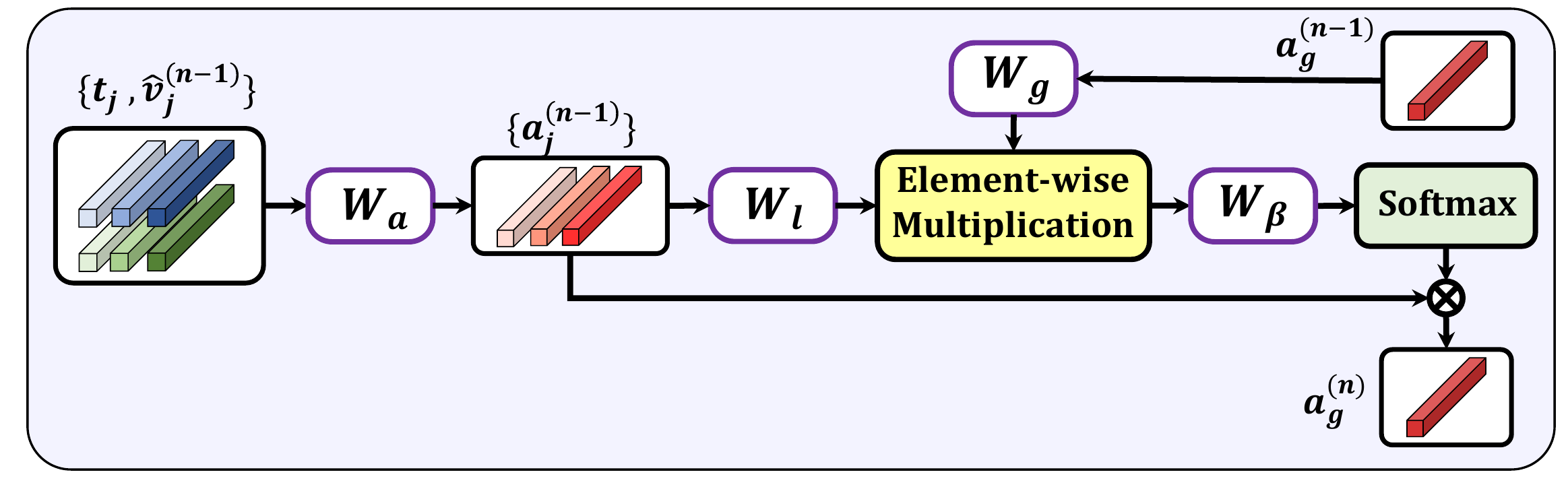} 
	\end{tabular}
	\caption{Illustration of RAR that updates the aggregation weights under the guidance of the holistic alignment vector in the previous step.}
	\label{fig:RAR}
\end{figure}

{\bf Discussion.} For word-region interactions, \textbf{1)} most existing approaches compute the one-time forward procedures with the fixed and uniform factors, which obviously lack the regulation ability to adapt itself to various words with diverse semantics. In contrast, the RCR first generates the constructed alignment that records the abundant correlation between each word and all related regions from the previous step, which in turn reweighs the weight vector and temperature value concerning each word to refine the corresponding attention distribution. 
\textbf{2)} Early works are always inclined to align the words with potentially "closest" regions in the comparable space even for negative image-text pairs. We assume that the words from positive pairs should focus more on specific and relevant regions, while the ones from negative pairs should attend to “completely irrelevant” regions. From the above perspective, the RCR can dynamically update the channel-wise measure and refine the numerical value of word-region relevance, thus leading to larger gaps between matched and unmatched pairs and greater capability in modeling complex matching patterns.

\begin{figure*}[t!]
	\centering
	\begin{tabular}{@{}c}
		\includegraphics[width=0.98\linewidth, height=0.24\linewidth,trim= 0 350 190 0,clip]{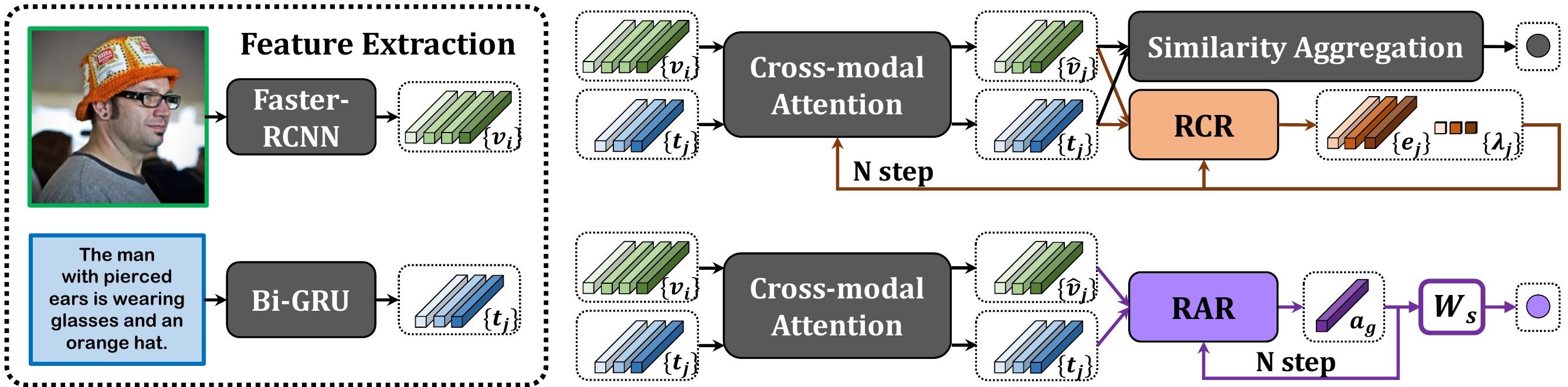} 
	\end{tabular}
	\caption{Illustration of plug-and-play operation with our regulators. For independent application, the RCR facilitates region-word correspondence and preserves the raw similarity calculation, while the RAR promotes more accurate similarity prediction and retains the original cross-modal interaction.}
	\label{fig:framework}
\end{figure*}

\subsection{Recurrent Aggregation Regulator}
The Recurrent Aggregation Regulator (RAR) aggregates the word-region alignments in a recurrent manner by progressively optimizing the aggregation weights guided by the holistic alignment at the early step in Fig.~\ref{fig:RAR}.
Given the word-attended alignment vector $\boldsymbol{a}_{j}$ in Eq.~\eqref{eq:aj}, we initialize a guidance alignment with
\begin{equation}
\label{eq:ag}
    \boldsymbol{a}_{g}^{(0)} = \frac{1}{L} \sum_{j=1}^{L} \boldsymbol{a}_{j} \ ,
\end{equation}
which actually performs the average pooling with the aggregation weight to be $1/L$ for each alignment.
Instead of directly using the averaged alignment $\boldsymbol{a}_{g}^{(0)}$ for inferring the similarity score, we iteratively update the aggregation weights under the guidance of $\boldsymbol{a}_{g}^{(0)}$ in the previous step:
\begin{equation}
\label{eq:betaj}
 \begin{split}
    \boldsymbol{u}_{j}^{(1)} &= tanh(\boldsymbol{W}_{g}\boldsymbol{a}_{g}^{(0)})\odot tanh(\boldsymbol{W}_{l}\boldsymbol{a}_{j}) \ , \\
    {\beta}_{j}^{(1)} &= \frac{exp(\boldsymbol{W}_{\beta}\boldsymbol{u}_{j}^{(1)})}{\sum_{i=1}^{L} exp(\boldsymbol{W}_{\beta}\boldsymbol{u}_{i}^{(1)})} \ ,
\end{split}
\end{equation}
where $\boldsymbol{W}_{g} \in \mathbb{R}^{m \times m}$, $\boldsymbol{W}_{l} \in \mathbb{R}^{m \times m}$, $\boldsymbol{W}_{\beta} \in \mathbb{R}^{m}$ are learnable parameters. The initial guidance alignment $\boldsymbol{a}_{g}^{(0)}$ in Eq.~\eqref{eq:ag} is then updated as follows:
\begin{equation}
\label{eq:updateag}
    \boldsymbol{a}_{g}^{(1)} = \sum_{j=1}^{L} {\beta}_{j}^{(1)} \boldsymbol{a}_{j} \ ,
\end{equation}
where ${\beta}_{j}^{(1)}$ is the updated aggregation weight for the $j$-th alignment.
For simplicity, we formulate the Recurrent Aggregation Regulator (RAR) as 
\begin{equation}
\label{eq:rar}
    \boldsymbol{a}_{g}^{(n)} = \mathbf{RAR}(\boldsymbol{a}_{g}^{(n-1)},\boldsymbol{A}) \ ,
\end{equation}
with $\boldsymbol{A} = \{\boldsymbol{a}_{1},...,\boldsymbol{a}_{L}\}$ indicating all the alignments constructed from the T2I attention as with Eq.~\eqref{eq:aj}.

The final similarity score can be inferred from $\boldsymbol{a}_{g}^{(n)}$ with a fully-connected layer as
\begin{equation}
\label{eq:st2iRAR}
    \mathcal{S}_{T2I}^{RAR} = sigmoid(\boldsymbol{W}_{s}\boldsymbol{a}_{g}^{(n)}) \ ,
\end{equation}
where $\boldsymbol{W}_{s} \in \mathbb{R}^{m}$ is a learnable parameter, and $sigmoid(\cdot)$ aims to output a similarity score within $[0,1]$.

{\bf Discussion.} Instead of averaging all the cosine similarities between all word features and attended image features as formulated in Eq.~\eqref{eq:st2i}, the RAR goes one step further by iteratively aggregating the constructed alignments to recognize more comprehensive contents across modalities. Specifically, the RAR starts from the average aggregation, and in each regulation step it attempts to learn from the contextual message outputs from the previous step and balance the importance of each word-based alignment without no manual tuning. It is observed that the RAR increasingly emphasizes more on the alignments from more significant words, and gradually reduces the aggregation weights from unimportant ones. By this means, the network constantly adjusts the proportion of all the alignments and assigns more plausible aggregation weights, resulting in a more discriminative holistic alignment and more appropriate distance metrics in image-text matching.

\subsection{Properties of RCR and RAR}
{\bf Plug-and-Play on Multiple Models.} The most attractive property of RCR and RAR is “plug-and-play”. To demonstrate their great applicability, we apply these two regulators to many existing methods based on cross-modal interaction:

\textit{Stacked Cross Attention
(SCAN)}~\cite{SCAN} first computes all region-word similarities and aligns each region/word with its corresponding words/regions. The final similarity is obtained by averaging all region/word-based cosine distances.

\textit{Bidirectional Focal Attention (BFAN)}~\cite{BFAN} extends the generic attention by reassigning more fine-grained attention weight for each region-word pair and calculates the matching result by summing up region-based and word-based scores.

\textit{Position Focused Attention
(PFAN)}~\cite{PFAN} enhances region features by introducing extra position information to promote region-word correspondences and integrates all region/word-attended cosine similarities as the prediction.

\textit{Cross-Modal Adaptive Message Passing
(CAMP)}~\cite{CAMP} explores a region-word affinity matrix via inner product and transfers cross-modality contents to improve the region and word representations, which are then aggregated as the holistic image and text features to compute the final similarity.

\textit{Similarity Graph Reasoning and Attention Filtration (SGRAF)}~\cite{SGRAF} adopts cosine similarities multiplied with a fixed temperature as region-word attention weights, followed by the complex graph and attention modules to map hierarchical similarity features into a matching score.

Fig.~\ref{fig:framework} illustrates how we plug the RCR or RAR into the above matching approaches. Specifically, cross-modal attention utilizes the cosine metric or inner product as region-word affinity weights, and outputs each region/word along with its related words/regions. With these paired features, the RCR first constructs the alignment vectors and then learns the corresponding weight vectors and temperature factors via Eq.~\eqref{eq:aj}-\eqref{eq:factors}, which in turn refine the region-word feature distances and optimize the cross-modal interaction via Eq.~\eqref{eq:updatecij}-\eqref{eq:updateahatvj}. Besides with a set of alignment vectors, the RAR progressively generates more appropriate weights between a guidance vector and all alignment vectors via Eq.~\eqref{eq:ag}-\eqref{eq:updateag}, and facilitates more rational similarity aggregation processing. It turns out that such simple message feedback brings remarkable improvements on many cross-modal interaction works, and even achieves superior performance than related complicated counterparts.

\begin{table*}[t!]
    \caption{Retrieval results in chronological order. The best two results are marked in \textbf{bold} and \underline{underline}. $\ast$ adopts warm-up strategy and text size augmentation, while $\star$ denotes ensemble models with high resolution of the input images.}
	\label{tabcocof30k}
	\centering
	\setlength{\tabcolsep}{1.15mm}{
	\begin{tabular}{l|cccccc|cccccc|cccccc}
		\hline
		\multirow{3}{*}{\bf Methods} 
		&\multicolumn{6}{c}{Flickr30K 1K Test} 
		&\multicolumn{6}{|c}{MSCOCO 5-fold 1K Test} 
		&\multicolumn{6}{|c}{MSCOCO 5K Test} \\
		\;  &\multicolumn{3}{c}{Sentence Retrieval} 
		&\multicolumn{3}{c}{Image Retrieval} &\multicolumn{3}{|c}{Sentence Retrieval} 
		&\multicolumn{3}{c}{Image Retrieval}
		&\multicolumn{3}{|c}{Sentence Retrieval} 
		&\multicolumn{3}{c}{Image Retrieval}\\
		\;  & R@1 & R@5 & R@10 & R@1 & R@5 & R@10 & R@1 & R@5 & R@10 & R@1 & R@5 & R@10 & R@1 & R@5 & R@10 & R@1 & R@5 & R@10\\
		\hline
		\multicolumn{19}{c}{\textbf{Faster-RCNN ( ResNet-101 BUTD~\cite{BU_TDA} ) + Random Word Embedding Initialization}} \\
		\hline
		SCAN\cite{SCAN}$_{\textit{ECCV18}}$ 
		&67.4 &90.3 &95.8 &48.6 &77.7 &85.2 
		&72.7 &94.8 &98.4 &58.8 &88.4 &94.8 
		&50.4 &82.2 &90.0 &38.6 &69.3 &80.4 \\
		VSRN\cite{VSRN}$_{\textit{ICCV19}}$ 
		&71.3 &90.6 &96.0 &54.7 &81.8 &88.2 
		&76.2 &94.8 &98.2 &62.8 &89.7 &95.1
		&53.0 &81.1 &89.4 &40.5 &70.6 &81.1\\
		CAAN\cite{CAAN}$_{\textit{CVPR20}}$ 
		&70.1 &91.6 &97.2 &52.8 &79.0 &87.9
		&75.5 &95.4 &98.5 &61.3 &89.7 &95.2
		&52.5 & 83.3 &90.9 &41.2 &70.3 &\textbf{82.9}\\
		IMRAM\cite{IMRAM}$_{\textit{CVPR20}}$ 
		&74.1 &93.0 &96.6 &53.9 &79.4 &87.2
		&76.7 &95.6 &98.5 &61.7 &89.1 &95.0
		&53.7 &83.2 &91.0 &39.7 &69.1 &79.8\\
		MMCA\cite{MMCA}$_{\textit{CVPR20}}$ 
		&74.2 &92.8 &96.4 &54.8 &81.4 &87.8
		&74.8 &95.6 &97.7 &61.6 &89.8 &95.2
		&54.0 &82.5 &90.7 &38.7 &69.7 &80.8\\
		GSMN\cite{GSMN}$_{\textit{CVPR20}}$
		&76.4 &\underline{94.3} &97.3 &57.4 &82.3 &\underline{89.0}
		&78.4 &\underline{96.4} &\underline{98.6} &\underline{63.3} &90.1 &95.7
		&-- &-- &-- &-- &-- &--\\
		SGRAF\cite{SGRAF}$_{\textit{AAAI21}}$
		&\underline{77.8} &94.1 &\underline{97.4} &\underline{58.5} &\underline{83.0} &88.8
		&\underline{79.6} &96.2 &98.5 &63.2 &\textbf{90.7} &\textbf{96.1}
		&\underline{57.8} &\underline{84.9} &\underline{91.6} &\underline{41.9} &\underline{70.7} &81.3\\
		SHAN\cite{SHAN}$_{\textit{IJCAI21}}$ 
		&74.6 &93.5 &96.9 &55.3 &81.3 &88.4
		&76.8 &96.3 &\textbf{98.7} &62.6 &89.6 &\underline{95.8} 
		&-- &-- &-- &-- &-- &--\\
		WCGL\cite{WCGL}$_{\textit{ICCV21}}$ 
		&74.8 &93.3 &96.8 &54.8 &80.6 &87.5
		&75.4 &95.5 &\underline{98.6} &60.8 &89.3 &95.3
		&-- &-- &-- &-- &-- &--\\
		{\bf RCAR(\cite{SCAN}\_T2I\;)}
		&\underline{77.8} &93.6 &96.9 &57.2 &82.8 &88.5
		&78.2 &96.3 &98.4 &62.2 &89.6 &95.3
		&57.4 &83.8 &91.0 &40.7 &69.8 &80.4\\
		{\bf RCAR(\cite{SCAN}\_I2T\;)}
		&74.7 &93.0 &97.1 &54.6 &80.5 &87.0
		&78.5 &95.9 &98.5 &61.2 &89.0 &95.2
		&56.6 &83.3 &91.2 &39.1 &68.7 &79.4\\
		{\bf RCAR(\cite{SCAN}\_All\;)} 
		&\textbf{78.7} &\textbf{94.6} &\textbf{97.6} &\textbf{59.5} &\textbf{84.0} &\textbf{89.5}
		&\textbf{80.6} &\textbf{96.6} &\underline{98.6} &\textbf{64.1} &\underline{90.5} &\underline{95.8} 
		&\textbf{59.6} &\textbf{85.8} &\textbf{92.4} &\textbf{42.5}  &\textbf{71.7} &\underline{81.8}\\
		\hline
		SCAN\cite{SCAN}$_{\textit{ECCV18}}{\ast}$
		&72.2 &92.4 &96.5 &53.6 &81.2 &88.6
		&72.8 &94.3 &98.0 &57.5 &87.8 &94.5 
		&50.1 &79.5 &88.1 &36.5 &66.7 &78.1\\
		GPO\cite{GPO}$_{\textit{CVPR21}}{\ast}$
		&78.0 &94.6 &97.8 &58.3 &84.6 &90.8
		&78.4 &96.2 &98.7 &62.7 &90.6 &95.9
		&56.8 &84.5 &91.4 &40.3 &70.7 &81.7\\
		GPO\cite{GPO}$_{\textit{CVPR21}}{\ast\star}$
		&\underline{80.7} &\textbf{96.4} &\underline{98.3} &\underline{60.8} &\textbf{86.3} &\textbf{92.3}
		&\underline{80.0} &\textbf{97.0} &\textbf{99.0} &\underline{64.8} &\textbf{91.6} &\textbf{96.5}
		&\underline{59.8} &\textbf{86.1} &\textbf{92.8} &\underline{42.7} &\underline{72.8} &\textbf{83.3}\\
		{\bf RCAR(\cite{SCAN}\_T2I$\ast$\;)}
		&79.7 &95.0 &97.4 &60.9 &84.4 &90.1
		&79.1 &96.5 &98.8 &63.9 &90.7 &95.9
		&59.1 &84.8 &91.8 &42.8 &71.5 &81.9\\
		{\bf RCAR(\cite{SCAN}\_I2T$\ast$\;)}
		&76.9 &95.5 &98.0 &58.8 &83.9 &89.3
		&79.3 &96.5 &98.8 &63.8 &90.4 &95.8 
		&58.4 &84.6 &91.9 &41.7 &71.4 &81.7\\
		{\bf RCAR(\cite{SCAN}\_All$\ast$\;)} 
		&\textbf{82.3} &\underline{96.0} &\textbf{98.4} &\textbf{62.6} &\underline{85.8} &\underline{91.1}
		&\textbf{80.9} &\underline{96.9} &\underline{98.9} &\textbf{65.7} &\underline{91.4} &\underline{96.4} 
		&\textbf{61.3} &\textbf{86.1} &\underline{92.6} &\textbf{44.3} &\textbf{73.2} &\underline{83.2}\\
		\hline
	\end{tabular}}
\end{table*}

\begin{algorithm}[t!]
    \caption{Cooperation of RCR and RAR (RCAR)}
    \label{alg:Framwork}
    \begin{algorithmic}[1]
    \Require
        Image features $\boldsymbol{V}$, text features $\boldsymbol{T}$, initial temperature $\lambda$ and weight vector $\mathbbm{1}^{d}$, and regulation steps $N$;
    \Ensure
        Final similarity score $\mathcal{S}^{RCAR}$; 
    \State Compute $\hat{\boldsymbol{v}}_{j}^{(0)}, j=1,...L$ with Eq.~\eqref{eq:cma};
    \State Compute $\boldsymbol{a}_j^{(0)}, j=1,...L$ with Eq.~\eqref{eq:aj};
    \State Compute $\boldsymbol{a}_g^{(0)}$ with Eq.~\eqref{eq:ag};
    \For{$n=1$ to $N$}
    \State Update $\hat{\boldsymbol{v}}_j^{(n)}, j=1,...L$ with Eq.~\eqref{eq:camrcr};
    \State Update $\boldsymbol{a}_j^{(n)}, j=1,...L$ with Eq.~\eqref{eq:aj};
    \State Update $\boldsymbol{a}_g^{(n)}$ with Eq.~\eqref{eq:rar};
    \EndFor
    \State Compute $\mathcal{S}^{RCAR}$ with Eq.~\eqref{eq:st2iRAR};\\
    \Return $\mathcal{S}^{RCAR}$
    \end{algorithmic}
\end{algorithm}

{\bf Cooperation of RCR and RAR.}
\label{secRCAR}
The RCR and RAR can cooperate with each other where the RCR is responsible for adjusting the cross-modal interaction and the RAR refines the alignment aggregation to achieve further improvements. In Algorithm~\ref{alg:Framwork}, we introduce an easy combination as RCAR that performs these two regulations one-by-one. Note that their cooperation is pretty flexible, and more variants with experimental results can be found in Sec.~\ref{secAS}.

\section{Experiments}
In this section, we first describe the detailed implementations and training settings, and then validate the great performance and generalization ability of two regulators.

\subsection{Datasets and Settings}
\textbf{Datasets and Protocols.} We utilize MSCOCO~\cite{MSCOCO} and Flickr30K~\cite{Flickr30k} that separately consist of 31,783 and 123,287 images, with each one annotated with 5 text descriptions. 
For Flickr30K, we split the dataset into 1,000 images for validation, 1,000 images for testing, and the rest for training.
For MSCOCO, we utilize 113,287 images for training, 5,000 images for validation, and 5,000 images for testing. We report the results by averaging over 5 folds of 1K test images and testing on the full 5K test images, respectively. In terms of evaluation metric, we measure the performance by the Recall@$\hat{K}$ (R@$\hat{K}$) which measures the fraction of queries whose ground-truth is ranked among the closest $\hat{K}$ results.

\textbf{Implementation Details.}
The bottom-up detector~\cite{BU_TDA} is used to generate the top $K$=36 region proposals with 2048 dimensions. Besides, we set the dimensions of word embedding, hidden state of BiGRU, and alignment vector as 300, $d$=1024, and $m$=256 respectively. The initial $\lambda^{(0)}$=10 and $e^{(0)}$=$\mathbbm{1}^{d}$ are updated by two MLPs of Input(256)-FC(128)-Tanh-FC(1) and Input(256)-FC(512)-Tanh-FC(1024)-Tanh.
The network is trained by the Adam optimizer~\cite{Adam} with a mini-batch size of 128. For MSCOCO, we set the learning rate to be 0.0002 for the first 10 epochs and 0.00002 for the next 10 epochs. For Flickr30K, the learning rate is set to be 0.0002 for 30 epochs and decayed by 0.1 for the next 10 epochs. 

\subsection{Quantitative Results and Analysis}
We present the results with $N$=2 RCAR (i.e. 2-step RAR and 1-step RCR) with the simplest SCAN and improved SCAN$\ast$ that adopts a warm-up strategy and text size augmentation as with~\cite{GPO}. We report the ensemble results of T2I and I2T models by averaging the individual scores offline.

\textbf{Results on Flickr30K.} 
TABLE~\ref{tabcocof30k} shows the retrieval results on Flickr30K. Compared with SCAN~\cite{SCAN}, our regulators can improve the absolute R@1 boost of $11.3\%$ and $10.9\%$ on sentence and image retrieval. Besides, the RCAR with the improved SCAN~\cite{SCAN}$\ast$ yields the bidirectional R@1 of $82.3\%$ and $62.6\%$ separately, and exceeds the best competitor GPO~\cite{GPO} by $4.3\%$ and $4.3\%$ under the same settings, indicating the significance of exploiting the regulation capabilities with adaptive correspondence and recurrent aggregation.

\textbf{Results on MSCOCO.} 
In TABLE~\ref{tabcocof30k} with 5-fold 1K test images, our RCAR can produce the state-of-the-art performance based on the simplest SCAN~\cite{SCAN}, and outweigh the SGRAF~\cite{SGRAF} by $1.0\%$ and $0.9\%$ on the most concerned R@1. Under the fair comparison, the improved version consistently surpasses the previous best method GPO~\cite{GPO} by $2.5\%$ and $3.0\%$ R@1 increases at two directions.
With the larger and more compelling 5K test images, our RCAR with~\cite{SCAN} and ~\cite{SCAN}$\ast$ can further outperform the SGRAF~\cite{SGRAF} and GPO~\cite{GPO} by 1.8/0.6$\%$ and 4.5/4.0$\%$ R@1 improvements respectively, validating the superior performance and generalization capability in handling more complex matching patterns. 

\begin{table}
    \caption{Retrieval results of plug-and-play RAR, RCR and RCAR on Flickr30K with the official codes of multiple approaches.}\label{tabMO}
    \centering
    \setlength{\tabcolsep}{1.18mm}{
    \begin{tabular}{l|cc|cccccc}
        \hline
        \multirow{2}{*}{\bf Methods}
        &\multirow{2}{*}{\#RAR}
        &\multirow{2}{*}{\#RCR}
        &\multicolumn{2}{c}{Sen. Ret.}
        &\multicolumn{2}{c}{Ima. Ret.}
        &Mem. &Tim.\\
        && &R@1&R@5
        &R@1&R@5 
        &($G$)&($us$)\\
        \hline
        \multirow{2}{*}{BFAN\cite{BFAN}} 
        &\xmark&\xmark
        &68.1&91.4 &50.8&78.4 &11.2 &11.8\\
        &3&\xmark &{\bf74.5}&{\bf92.9}&{\bf54.5}&{\bf80.6} &12.4 &14.4\\
        \hline
        \multirow{2}{*}{SGRAF\cite{SGRAF}} 
        &\xmark&\xmark 
        &77.8&94.1&58.5&83.0 &12.4 &9.4\\
        &\xmark&2 &{\bf79.2}&{\bf94.3}&{\bf59.7}&{\bf83.1} &13.4 &28.4\\
        \hline
        \multirow{4}{*}{CAMP\cite{CAMP}} 
        &\xmark&\xmark 
        &68.1&89.7&51.5&77.1 &18.8 &36.8\\
        &\xmark&2 
        &75.1&93.2&56.0&81.3 &20.1 &78.5\\
        &3&\xmark 
        &74.4&91.9 &53.7&80.0 &19.0 &43.1\\
        &2&1 
        &{\bf76.3}&{\bf93.3}&{\bf57.1}&{\bf81.6} &19.4 &55.3\\
        \hline
        \multirow{4}{*}{PFAN\cite{PFAN}} 
        &\xmark&\xmark 
        &70.0&91.8&50.4&78.7 &10.4 &8.8\\
        &\xmark&2 
        &73.6&92.5&54.3&81.1 &11.7 &43.0\\
        &3&\xmark 
        &78.1&94.1&58.4&82.9 &10.8 &12.6\\
        &2&1 
        &{\bf80.1}&{\bf95.7}&{\bf59.9}&{\bf84.4} &10.9 &26.1\\
        \hline
    \end{tabular}}
\end{table}

\textbf{Plug-and-Play on Multiple Models.} 
We attempt to apply our regulators on a series of representative works including BFAN~\cite{BFAN}, SGRAF~\cite{SGRAF}, CAMP~\cite{CAMP}, and PFAN~\cite{PFAN} on Flickr30K in TABLE~\ref{tabMO}.
\textbf{1)} Since \textbf{BFAN}~\cite{BFAN} designs specific cross-modal attention which explores a novel bidirectional focal attention to eliminate irrelevant fragments from the shared semantic, we just plug 3-step RAR into the network and obtain the R@1 gains of $6.4\%$ and $3.7\%$ on BFAN. Note that when both applied with the RAR, the cross-modal attention unit from SCAN~\cite{SCAN} refined by the RCR achieves much better performance (R@1=$78.7/59.5\%$) than BFAN (R@1=$74.5/54.5\%$), which further verifies the superior cross-modal correspondence by exploiting the regulation abilities of the network itself.
\textbf{2)} \textbf{SGRAF}~\cite{SGRAF} employs graph reasoning and attention filtration to refine the cross-modal representations, but ignores the ability of cross-modal attention unit, which can work with the RCR to empower flexible region-word interactions. With 2-step RCR, SGRAF obtains a maximum $1.4\%$ increase on R@1, reflecting general effectiveness with the complicated network.
\textbf{3)} \textbf{CAMP}~\cite{CAMP} and \textbf{PFAN}~\cite{PFAN} integrate the cross-modal message flow and valuable position embedding separately to enhance the multi-modal representations, which can possess more powerful cross-modal interaction and aggregation actuated by our regulators. When the RCR/RAR/RCAR is introduced, the bidirectional R@1 can rise by $7.0$/$6.3$/$8.2\%$ and $4.5$/$2.2$/$5.6\%$ on CAMP, as well as $3.6$/$8.1$/$10.1\%$ and $3.9$/$8.0$/$9.5\%$ on PFAN on sentence and image retrieval respectively, demonstrating the strong compatibility and flexibility of our approach. 
\textbf{4) Computational cost.} Here, we report the memory and time consumption for prediction and average the additional cost for ensemble models. With 3-step RAR, the extra time increase of each image-text pair for BFAN/CAMP/PFAN is 2.6/6.3/3.8 $us$ with the memory increase of 1.2/0.2/0.4 $G$, while with 2-step RCR, the extra cost for SGRAF/CAMP/PFAN of 19/41.7/34.2 $us$ and 1.0/1.3/1.3 $G$. Besides, our RCAR brings the time and memory cost for CAMP/PFAN of 18.5/17.3 $us$ and 0.6/0.5 $G$, and gains a good balance between accuracy and complexity.

\subsection{Ablation Studies}
\label{secAS}
In this section, we first report the configurations of our proposed regulators, as well as the initialization and optimization of the attention factors. Then, we delve into the RAR and RCR to display how the aggregation weights and cross-attention distributions are progressively refined. Finally, we also explore alternative strategies and architectures. All comparisons are implemented based on SCAN~\cite{SCAN} unless otherwise noted.

\begin{table}[t!]
	\centering
	\caption{Residual design of the regulators with T2I attention on Flickr30K. Res denotes whether to use residual connection.}\label{tabstRCAR}
	 \setlength{\tabcolsep}{1.7mm}{
		\begin{tabular}{c|cc|cccccc}
			\hline
			\multirow{2}{*}{Model}
			&\multirow{2}{*}{Res}&\multirow{2}{*}{Step}
			&\multicolumn{3}{c}{Sentence Retrieval}
			&\multicolumn{3}{c}{Image Retrieval}\\
			&& &R@1&R@5&R@10&R@1&R@5&R@10\\
			\hline 
			Baseline &\xmark&\xmark &64.8 &89.9 &94.5 &46.9 &76.0 &84.5\\
			\hline
			\multirow{4}{*}{RCR} 
			&\xmark&1 &69.9 &91.1& 95.4 &51.9 &79.9 &86.0\\
			&\xmark&2 &61.4 &85.1 &91.2 &43.0 &72.5 &81.7\\
			&\cmark&1 &73.0 &\textbf{93.3} &\bf{97.4} &55.3 &80.2 &86.9\\
			&\cmark&2 &\bf{74.3} &\textbf{93.3} &97.1 &\bf{56.6} &\textbf{81.3} &\bf{87.8}\\
			\hline
			\multirow{4}{*}{RAR} 
			&\xmark&2 &75.7&93.6&\bf{97.2}&56.0&\textbf{81.8}&88.0\\
			&\xmark&3 &76.2&\textbf{93.8}&96.8&56.7&\textbf{81.8}&\bf{88.2}\\
			&\cmark&2 &\bf{76.6} &93.4 &96.5 &\bf{56.8} &81.2 &86.9\\
			&\cmark&3 &75.8 &92.9 &96.7 &56.6 &80.6 &85.8\\
			\hline
		\end{tabular}}
\end{table}

\textbf{Residual mechanism of the regulators.} 
In TABLE~\ref{tabstRCAR}, we carry out critical analyses of the influence of residual architectures. The Baseline employs the T2I attention from SCAN~\cite{SCAN} and averages all the cosine similarities as the final score. 
\textbf{1) Correspondence regulator.} Eq.~\eqref{eq:factors} indicates that the current adaptive weight vector $\boldsymbol{e}_{j}^{(n)}$ and softmax temperature $\lambda_{j}^{(n)}$ require the $\boldsymbol{e}_{j}^{(n-1)}$ and $\lambda_{j}^{(n-1)}$ at the last step. Here, we remove these two variables to construct a no-residual version of the RCR.
Compared with the residual structure, the RCR without residual design results in an obvious R@1 drop in TABLE~\ref{tabstRCAR}, indicating that RCR is inclined to predict offsets against the current state to adjust previous regulation dynamically. To be specific, 1-step RCR without residual fashion produces better results than Baseline. This is because in the beginning, each word shares the same initialization of a weight vector $\mathbf{e}^{(0)}$=$\mathbbm{1}^{d}$ and temperature $\lambda^{(0)}$=$10$, and the RCR barely infers the absolute value of these attention factors in the next step. However, after a 1-step adjustment, all the word-region interactions start from very distinct conditions (aligned or not) and attention states with regard to the particular words, making it difficult to further forecast absolute valuations. Therefore, the RCR with residual mechanism can better adjust the dynamic learning process and reduce the burden of one-time total optimization in a progressive manner. 
\textbf{2) Aggregation regulator.} Eq.~\eqref{eq:updateag} denotes that the current guidance alignment $\boldsymbol{a}_{g}^{(n)}$ requires no need for the $\boldsymbol{a}_{g}^{(n-1)}$ in the last iteration. Similarly, we average the early and learned alignments as the current guidance vector to build a residual version of the RAR.
The RAR aims to construct better bootstrap guidance and assign appropriate aggregation weights with the original word-based alignments throughout the process. Therefore, we can discover that the RAR with residual structure fails to bring significant improvements with the same word-attended alignments at each step. 

\begin{table}[t!]
	\caption{Impact of \#RCR and \#RAR with T2I attention on MSCOCO5K. Limited by the machine, we set the maximum step of RCR to 4.}
	\label{tabRCAR-T2I-COCO5K}
	\centering
	    \setlength{\tabcolsep}{1.8mm}{
		\begin{tabular}{cc|cccccc}
			\hline
			\multirow{2}{*}{\#RAR}&\multirow{2}{*}{\#RCR}
			&\multicolumn{3}{c}{Sentence Retrieval}
			&\multicolumn{3}{c}{Image Retrieval}\\
			&&R@1&R@5&R@10&R@1&R@5&R@10\\
			\hline 
			\xmark&\xmark &44.0 &75.5 &85.5 &32.6 &62.0 &74.4\\
			\hline
			1&\xmark &56.3 &82.9 &89.8 &39.6 &69.2 &79.8\\
			2&\xmark &56.2 &83.1 &90.7 &40.0 &\textbf{69.4} &79.9\\
			3&\xmark &\bf{56.6} &\textbf{83.5} &90.8 &\bf{40.5} &\textbf{69.4} &\bf{80.4}\\
    		4&\xmark &\bf{56.6} &83.3 &\bf{90.9} &40.4 &\textbf{69.4} &80.2\\
			\hline
			\xmark&1 &47.8 &79.4 &89.2 &35.2 &66.8 &78.4\\
			\xmark&2 &\bf{53.4} &81.9 &90.4 &\bf{38.4} &68.7 &79.9\\
		    \xmark&3 &52.4 &81.8 &90.6 &38.3 &69.0 &\bf{80.0}\\
		    \xmark&4 &53.1 &\textbf{83.3} &\bf{90.9} &38.3 &\textbf{69.3} &79.8\\
		    \hline
		    2&1 &\bf{57.4} &\textbf{83.8} &\bf{91.0} &40.7 &69.8 &80.4\\
		    3&2 &56.8 &\textbf{83.8} &\textbf{91.0} &\bf{40.8} &\textbf{70.0} &\bf{80.5}\\
			\hline
		\end{tabular}}
\end{table}

\begin{table}[t!]
	\caption{Impact of \#RCR and \#RAR with I2T attention on MSCOCO5K. Limited by the machine, we set the maximum step of RCR to 3.}
	\label{tabRCAR-I2T-COCO5K}
	\centering
	    \setlength{\tabcolsep}{1.8mm}{
		\begin{tabular}{cc|cccccc}
			\hline
			\multirow{2}{*}{\#RAR}&\multirow{2}{*}{\#RCR}
			&\multicolumn{3}{c}{Sentence Retrieval}
			&\multicolumn{3}{c}{Image Retrieval}\\
			&&R@1&R@5&R@10&R@1&R@5&R@10\\
			\hline 
			\xmark&\xmark &43.4 &74.8 &84.8 &32.0 &61.8 &74.2\\
			\hline
			1&\xmark &52.7 &81.6 &90.2 &36.8 &68.0 &78.8\\
			2&\xmark &\textbf{53.3} &\textbf{82.1} &\textbf{90.5} &38.1 &68.4 &79.2\\
			3&\xmark &53.1 &81.8 &90.4 &\textbf{38.4} &\textbf{68.6} &\textbf{79.3}\\
			\hline
			\xmark&1 &48.5 &80.5 &89.4 &34.9 &66.7 &78.5\\
			\xmark&2 &54.2 &82.0 &90.3 &38.3 &67.8 &78.9\\
		    \xmark&3 &\textbf{56.8} &\textbf{83.2} &\textbf{91.0} &\textbf{39.4} &\textbf{68.8} &\textbf{79.3}\\
		    \hline
		    2&1 &\textbf{56.6} &\textbf{83.3 }&91.2 &39.1 &68.7 &\textbf{79.4}\\
		    3&2 &56.0 &83.2 &\textbf{91.3} &\textbf{39.3} &\textbf{69.1} &\textbf{79.4}\\
			\hline
		\end{tabular}}
\end{table}

\textbf{Hyperparameter tuning of \#RAR and \#RCR.}
TABLE~\ref{tabRCAR-T2I-COCO5K}, \ref{tabRCAR-I2T-COCO5K}, \ref{tabRCAR-T2I-F30K} and \ref{tabRCAR-I2T-F30K} demonstrate the evaluation results of different steps about our regulators. We establish the baseline without any regulator that only utilizes the cross-modal attention~\cite{SCAN} and predicts the final score by averaging all the cosine distances via Eq.~\eqref{eq:st2i}.
\textbf{1) Aggregation regulator.} The RAR holds the original cross-modal attention unit and calculates the similarity by Eq.~\eqref{eq:st2iRAR} with the alignments constructed from Eq.~\eqref{eq:aj}. For MSCOCO 5K test set, the RAR can steadily improve the R@1 on sentence and image retrieval by at most 12.6\% and 7.9\% based on T2I attention, as well as 9.9\% and 6.4\% based on I2T attention. For Flickr30K 1k test set, it can also boost the bidirectional R@1 with consistent gains of over 9.8/7.5\% and 2.4/7.7\% upon T2I and I2T attention, respectively. We can see that the RAR can generate more accurate and plausible image-text similarity measurements. 
\begin{table}[t!]
	\caption{Impact of \#RCR and \#RAR with T2I attention on Flickr30K. Limited by the machine, we set the maximum step of RCR to 4.}
	\label{tabRCAR-T2I-F30K}
	\centering
	    \setlength{\tabcolsep}{1.8mm}{
		\begin{tabular}{cc|cccccc}
			\hline
			\multirow{2}{*}{\#RAR}
			&\multirow{2}{*}{\#RCR}
			&\multicolumn{3}{c}{Sentence Retrieval}
			&\multicolumn{3}{c}{Image Retrieval}\\
			&&R@1&R@5&R@10&R@1&R@5&R@10\\
			\hline 
			\xmark&\xmark &64.8 &89.9 &94.5 &46.9 &76.0 &84.5\\
			\hline
			1&\xmark &74.6&92.6&96.1&54.4&80.4&87.2\\
			2&\xmark &75.7&93.6&\textbf{97.2}&56.0&\textbf{81.8}&88.0\\
			3&\xmark &\textbf{76.2}&\textbf{93.8}&96.8&\textbf{56.7}&\textbf{81.8}&\textbf{88.2}\\
			4&\xmark &74.8&92.7&96.8&56.3&81.5&86.6 \\
			\hline
			\xmark&1 &73.0&93.3&97.4&55.3&80.2&86.9\\
			\xmark&2 &74.3&93.3&97.1&56.6&81.3&\textbf{87.8}\\
			\xmark&3 &73.9&93.1&96.4&56.0&81.4&87.5\\
			\xmark&4 &\textbf{75.2}&\textbf{94.1}&\bf{97.8}&\textbf{56.8}&\textbf{81.8}&\textbf{87.8}\\
			\hline
		    2&1&\bf{77.8}&93.6&96.9&\bf{57.2}&82.8&\bf{88.5}\\
			3&2&76.8&\textbf{94.1}&97.1&57.1&\textbf{83.0}&88.2\\
			\hline
		\end{tabular}}
\end{table}
\begin{table}[t!]
	\caption{Impact of \#RCR and \#RAR with I2T attention on Flickr30K. Limited by the machine, we set the maximum step of RCR to 3.}
	\label{tabRCAR-I2T-F30K}
	\centering
	    \setlength{\tabcolsep}{1.8mm}{
		\begin{tabular}{cc|cccccc}
			\hline
			\multirow{2}{*}{\#RAR}
			&\multirow{2}{*}{\#RCR}
			&\multicolumn{3}{c}{Sentence Retrieval}
			&\multicolumn{3}{c}{Image Retrieval}\\
			&&R@1&R@5&R@10&R@1&R@5&R@10\\
			\hline 
			\xmark&\xmark &66.7 &89.1 &94.0 &41.2 &72.7 &81.8\\
			\hline
			1&\xmark &69.1 &\textbf{91.2} &95.6 &48.9 &77.1 &85.1\\
			2&\xmark &\bf{70.8} &90.3 &\bf{95.7} &\bf{51.9} &\textbf{78.5} &\bf{85.9}\\
			3&\xmark &70.6 &\textbf{91.2} &95.0 &50.6 &77.8 &85.3\\
			\hline
			\xmark&1 &69.0 &\textbf{93.3} &96.8 &53.0 &80.4 &86.2\\
			\xmark&2 &73.3 &92.3 &96.7 &54.3 &80.5 &87.3\\
			\xmark&3 &\bf{75.7} &93.0 &\bf{97.4} &\bf{56.8} &\textbf{81.6} &\bf{88.1}\\
			\hline
		    2&1 &\bf{74.7} &\textbf{93.0} &\bf{97.1} &54.6 &80.5 &\bf{87.0}\\
		    3&2 &74.2 &92.4 &96.5 &\bf{54.7} &\textbf{80.6} &86.9\\
			\hline
		\end{tabular}}
\end{table}
\textbf{2) Correspondence regulator.} The RCR renews the region-word interactions to update aggregated features targeting the cross-modality instances iteratively, and keeps the raw prediction process through Eq.~\eqref{eq:st2i}. Compared with the foundation models, the RCR can obtain the steady R@1 increases by maximum 9.4/5.8\% (T2I) and 13.4/7.4\% (I2T) on MSCOCO5K, and meanwhile 10.4/9.9\% (T2I) and 9.0/15.6\% (I2T) on Flickr30K, verifying that RCR is capable of exploiting more fine-grained and appropriate word-region associations. 
\textbf{3) Cooperative regulators.} We employ the one-by-one combination of RAR and RCR as described in Algorithm~\ref{alg:Framwork}, indicating that the former always takes one more step than the latter. Compared with 2-step RAR and 1-step RCR, $N$=2 RCAR can further promote the R@1 at two directions by a large margin, demonstrating the good compatibility between RCR and RAR. Actually, their cooperations are pretty flexible. To take T2I attention on Flickr30K in TABLE~\ref{tabRCAR-T2I-F30K} as an example, an alternative strategy is to first perform 2-step RCR followed by 3-step RAR, which yields the competitive 77.5 and 57.8\% R@1 (against 77.8 and 57.2\% R@1 according to Algorithm~\ref{alg:Framwork}) on sentence and image retrieval separately.
Besides, we can also observe that larger $\#$RCR and $\#$RAR are not necessarily better, which may be due to the recurrent structure where a certain number of steps can saturate the performance of the network. 
\textbf{4) Computational cost.} The model size of single SCAN~\cite{SCAN} is 12.2$M$, and each step of RAR, RCR, or RCAR brings the extra parameters of nearly 0.13$M$, 0.95$M$, or 1.12$M$.
Using NVIDIA GeForce RTX 3090, the inference time of T2I-SCAN is 3.05 $us$ for each image-text pair with an extra cost of 0.51/4.72/5.91 $us$ per step by RAR/RCR/RCAR, while the predicted time of I2T-SCAN is 4.25 $us$ with an additional cost of 1.19/8.21/10.54 $us$.
From these experiments, we suggest step=2-3 for independent application and step=2(RAR)+1(RCR) for their cooperation, as they achieve a better trade-off between accuracy and complexity, and have proved general effectiveness and broad applicability on multiple state-of-the-art approaches in TABLE~\ref{tabMO}.

\begin{figure}[htpb]
	\centering
	\begin{tabular}{@{}c}
    	\includegraphics[width=0.94\linewidth, height=0.5\linewidth,trim=58 280 80 320,clip]{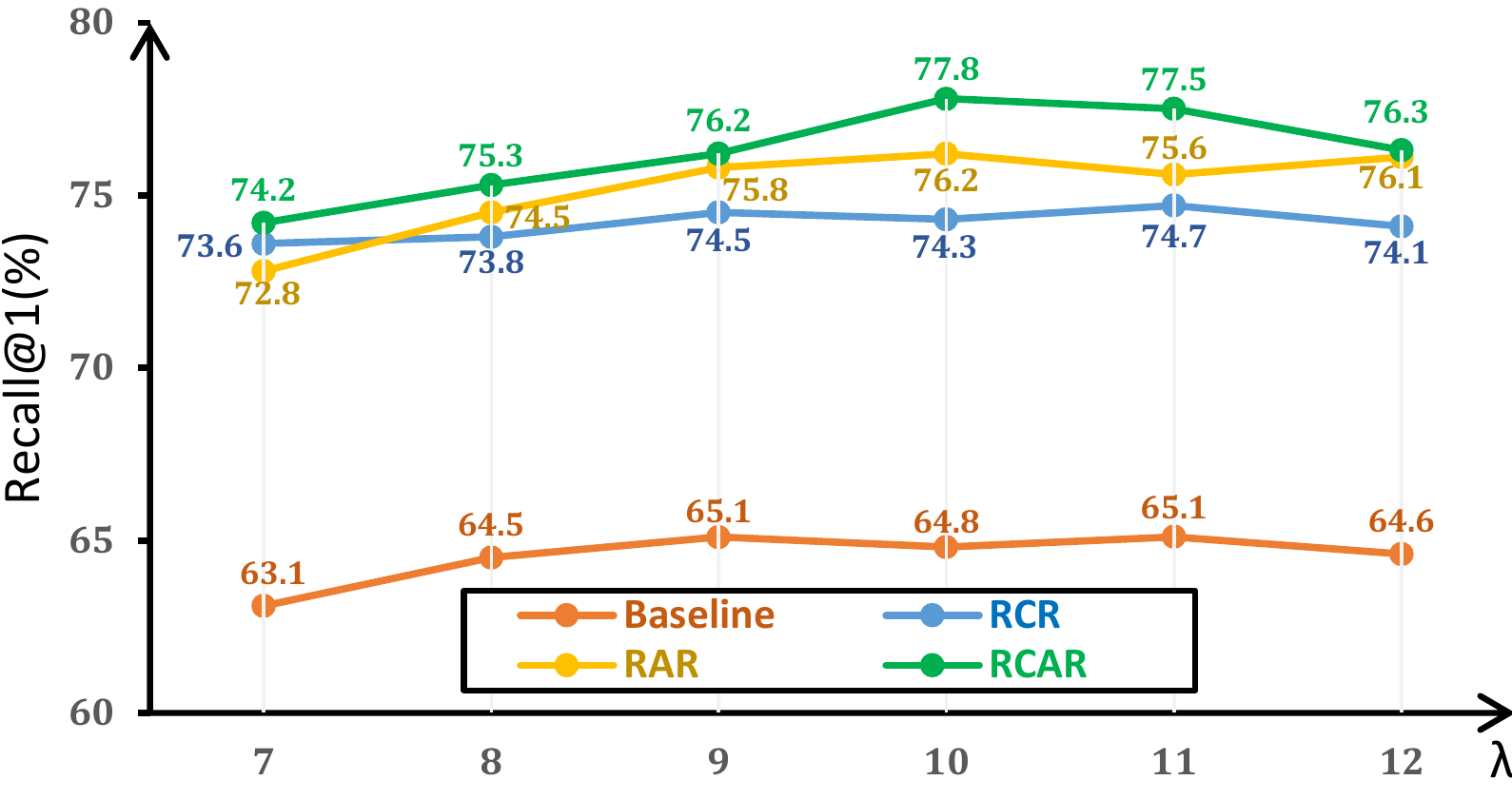} 
	\end{tabular}
	\caption{Impact of the $\lambda$ initialization with T2I attention on Flickr30K. The step of RAR, RCR and RCAR is set as 3, 2, and 2, respectively.}
	\label{fig:lambda}
\end{figure}

\begin{table}
	\centering
	\caption{Impact of the attention factors ($\lambda,\boldsymbol{e}$) optimization with T2I attention on Flickr30K. We adopt $N$=2 RCAR as a reference.}\label{tabRCRfeedback}
		\begin{tabular}{l|cccccc}
			\hline
			\multirow{2}{*}{Model}
			&\multicolumn{3}{c}{Sentence Retrieval}
			&\multicolumn{3}{c}{Image Retrieval}\\
			&R@1&R@5&R@10&R@1&R@5&R@10\\
			\hline 
			Fixed &75.7&93.6&\bf{97.2}&56.0&81.8&88.0\\
			Learnable &76.3&\bf{94.1}&96.8&56.9&81.3&87.3\\
			MLP$(\boldsymbol{t}_j)$ &75.9&93.3&96.5&56.6&81.2&87.3\\
		    MLP$(\boldsymbol{a}_j)$ &\bf{77.8}&93.6&96.9&{\bf57.2}&\bf{82.8}&\bf{88.5}\\
			\hline
		\end{tabular}
\end{table}

\textbf{Initialization of the attention factors.}
Fig.~\ref{fig:lambda} depicts the Recall@1(\%) at sentence retrieval with varying $\lambda$ value on Flickr30K. The $\lambda$ determines the initial distributions of word-region interactions where a large one tends to retain only the highly correlated instances while a small one results in the interference from irrelevant instances. Here, we take T2I attention as an example and compare the performance variation with $\lambda$ ranging from 7 to 12. We can see that our regulators can obtain the maximum performance benefit when $\lambda$=10, and achieve a consistent improvement among various settings, confirming the robustness and stability of our proposed method. It is worth noting that for simplicity and fair comparison, we directly set $\mathbf{e}^{(0)}$=$\mathbbm{1}^{d}$ and use exclusive $\lambda$ of each work as $\lambda^{(0)}$ in TABLE~\ref{tabMO}, which attempts to maintain the appropriate initialization of the incipient cross-attention unit.

\textbf{Optimization of the attention factors.} We investigate the different update strategies with $N$=2 RCAR in TABLE~\ref{tabRCRfeedback}. \textbf{1) Fixed}: We set the weight vector $\boldsymbol{e}$=$\mathbbm{1}^{d}$ and the softmax temperature $\lambda$=$10$ in the whole process; \textbf{2) Learnable}: The parameters $\boldsymbol{e}$ and $\lambda$ are learnable during the training, with initialization of $\mathbbm{1}^{d}$ and $10$ in the beginning; \textbf{3) MLP$(\boldsymbol{t}_j)$}: The attention factors of each word are learned with the original word features; \textbf{4) MLP$(\boldsymbol{a}_{j})$}: The attention factors of each word are learned with the constructed alignment vector. Note that \textbf{Learnable} achieves slightly better performance than \textbf{Fixed}, and adjusts the $\lambda=11.23$ for maximum performance benefit in experiments. However, a common problem of the two methods is the lack of capability to refine parameters adaptively to handle the diversity of different words.
\textbf{MLP$(\boldsymbol{t}_{j})$} produces even worse results than \textbf{Learnable}, which may be due to the lack of word-image alignment information, leading to difficulties in learning reasonable attention factors. In comparison, \textbf{MLP$(\boldsymbol{a}_{j})$} achieves the best performance by learning adaptive factors for each word, indicating the significance of exploiting the interaction feedback for better regulation.

\begin{figure}[t!]
	\centering
	\begin{tabular}{@{}c}
		\includegraphics[width=0.99\linewidth, height=0.86\linewidth, trim= 0 0 70 10, clip]{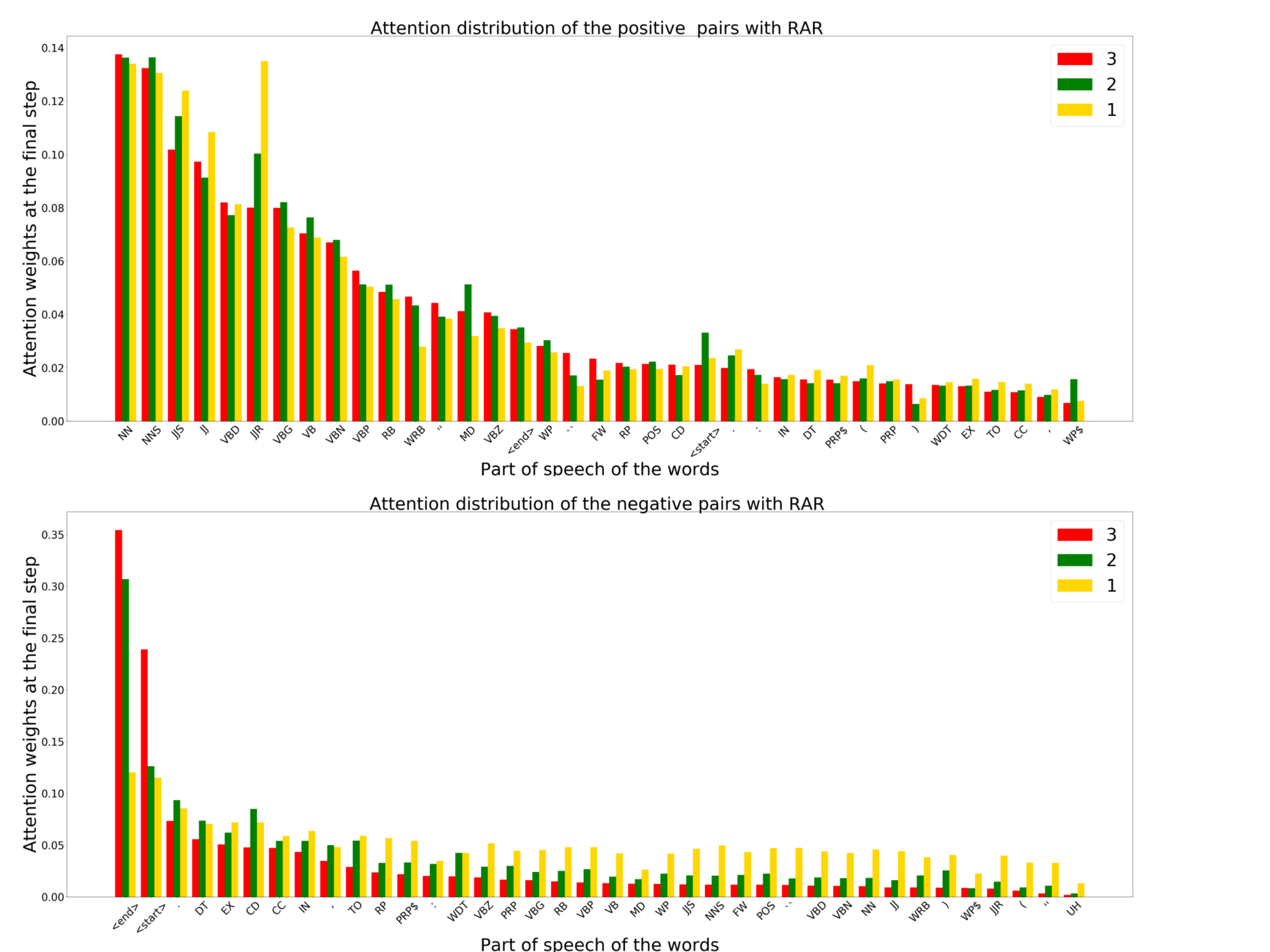} 
	\end{tabular}
	\caption{Quantitative weight statistics of word-attended alignments by n-step RAR with T2I attention on Flickr30K. 1, 2, 3 indicate the steps of the RAR. The top and bottom represent positive and negative image-text pairs.}
	\label{fig:RARattn}
\end{figure}

\begin{figure}[t!]
	\centering
	\begin{tabular}{@{}c}
		\includegraphics[width=0.99\linewidth, height=0.41\linewidth, trim= 0 133 150 10, clip]{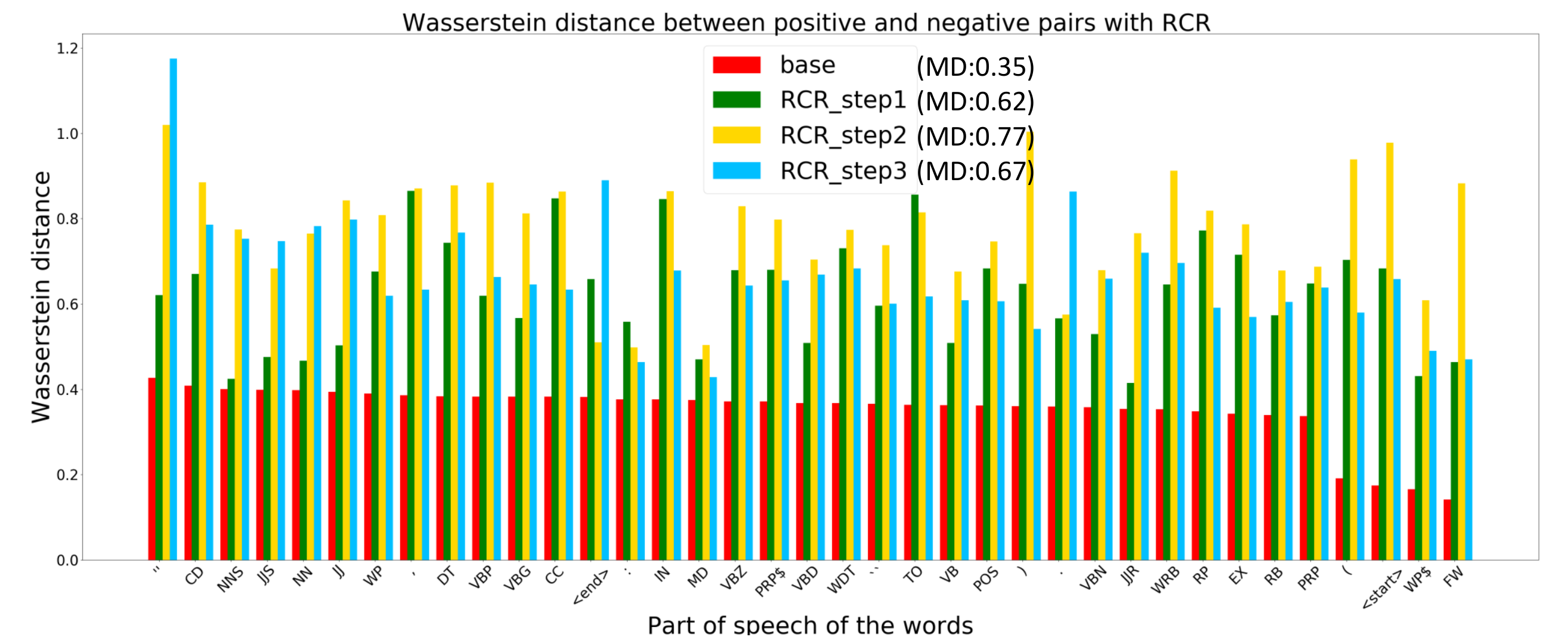} 
	\end{tabular}
	\caption{Quantitative distance statistics of word-attended cosine similarities by n-step RCR with T2I attention on Flickr30K. Base denotes T2I-SCAN~\cite{SCAN}. MD denotes the mean wasserstein distance with respect to all parts of speech.}
	\label{fig:RCRdistance}
\end{figure}

\begin{figure*}[ht]
	\centering
	\begin{tabular}{@{}cc}
		\includegraphics[width=0.49\linewidth, height=0.32\linewidth, trim=0 0 170 0,clip]{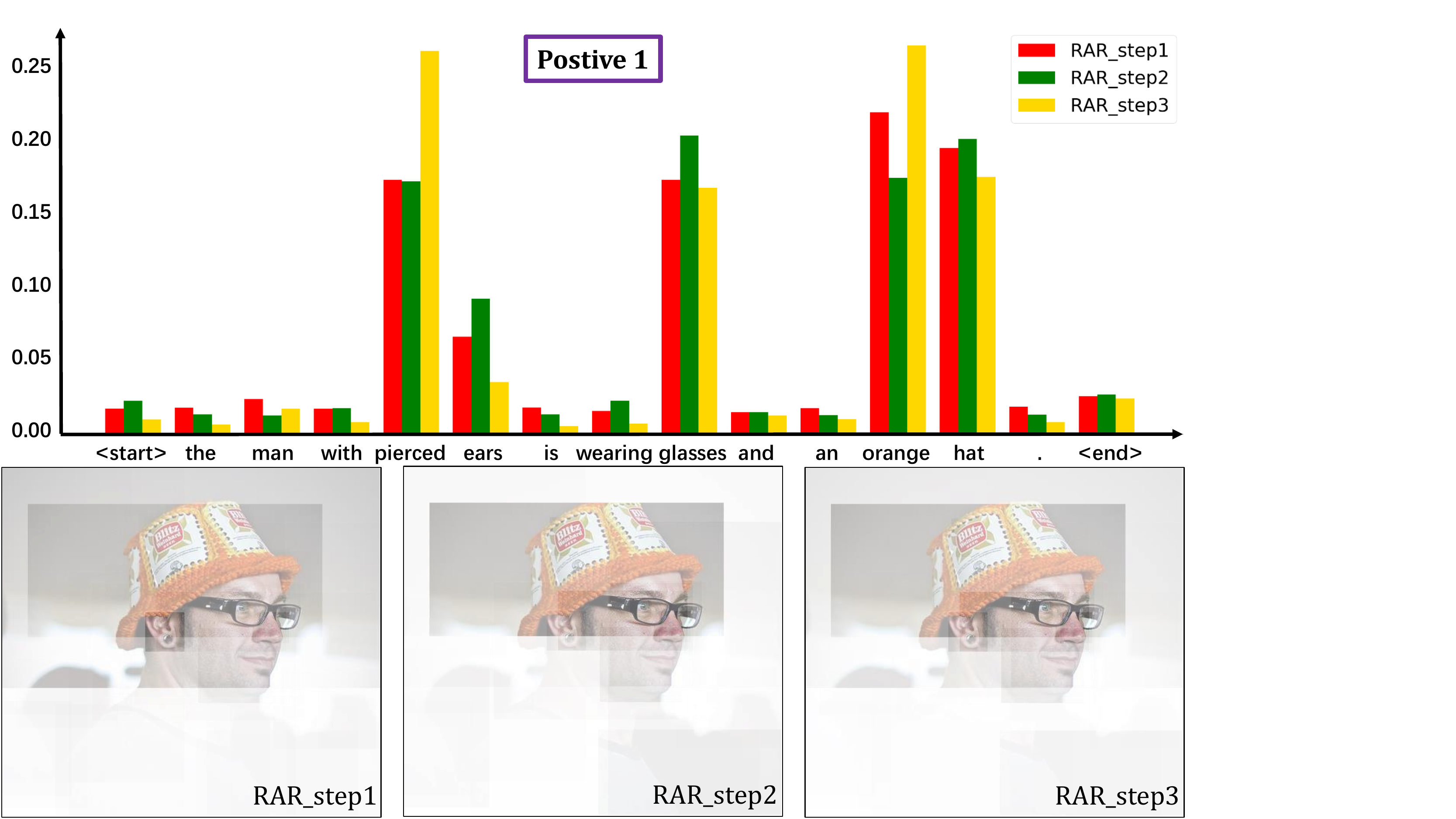}
		&\includegraphics[width=0.49\linewidth, height=0.32\linewidth, trim=0 60 170 0,clip]{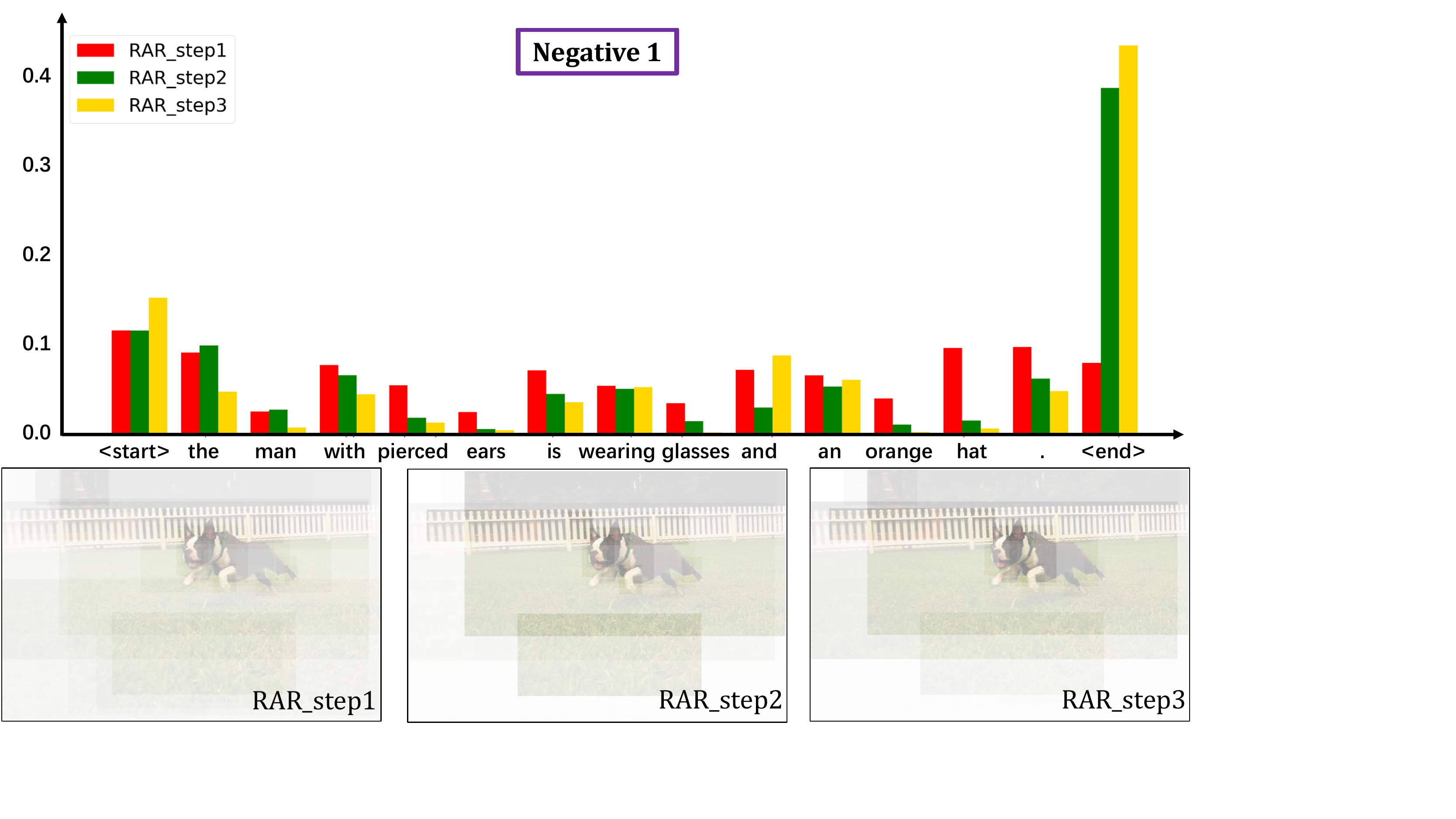} \\
		\includegraphics[width=0.49\linewidth, height=0.32\linewidth, trim=0 60 170 0,clip]{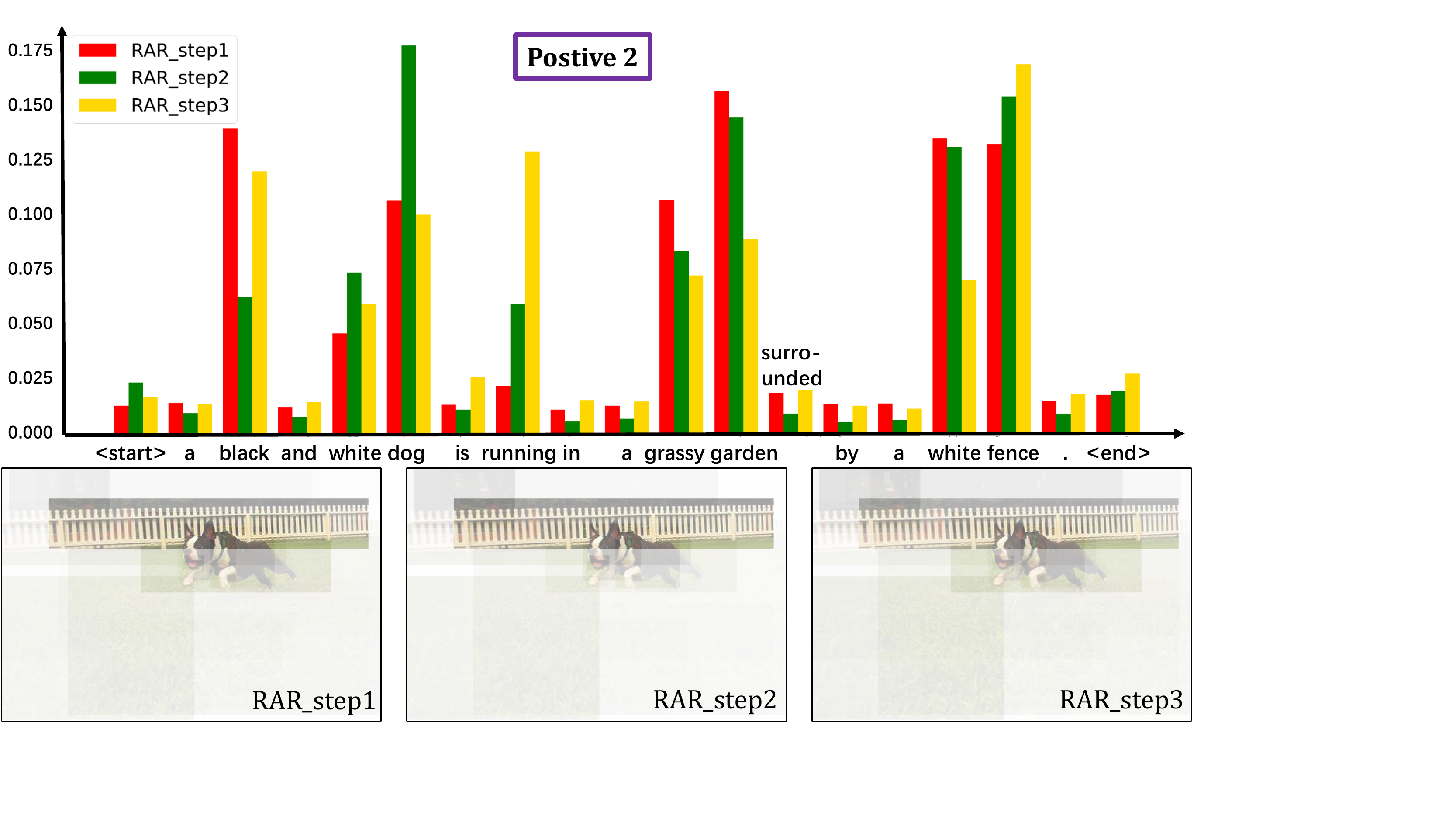}
		&\includegraphics[width=0.49\linewidth, height=0.32\linewidth, trim=0 0 170 0,clip]{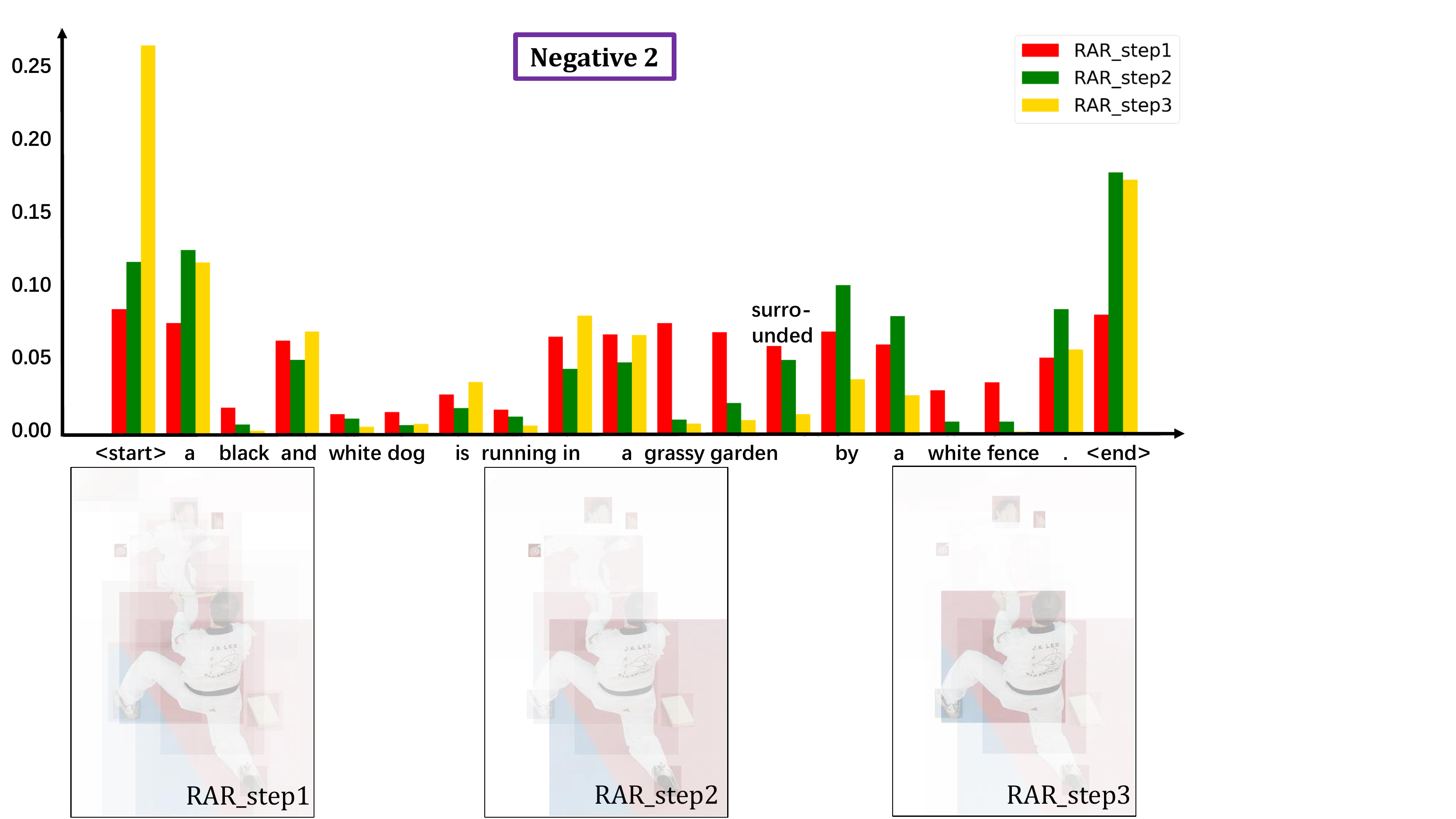} \\
		\includegraphics[width=0.49\linewidth, height=0.32\linewidth, trim=0 0 170 0,clip]{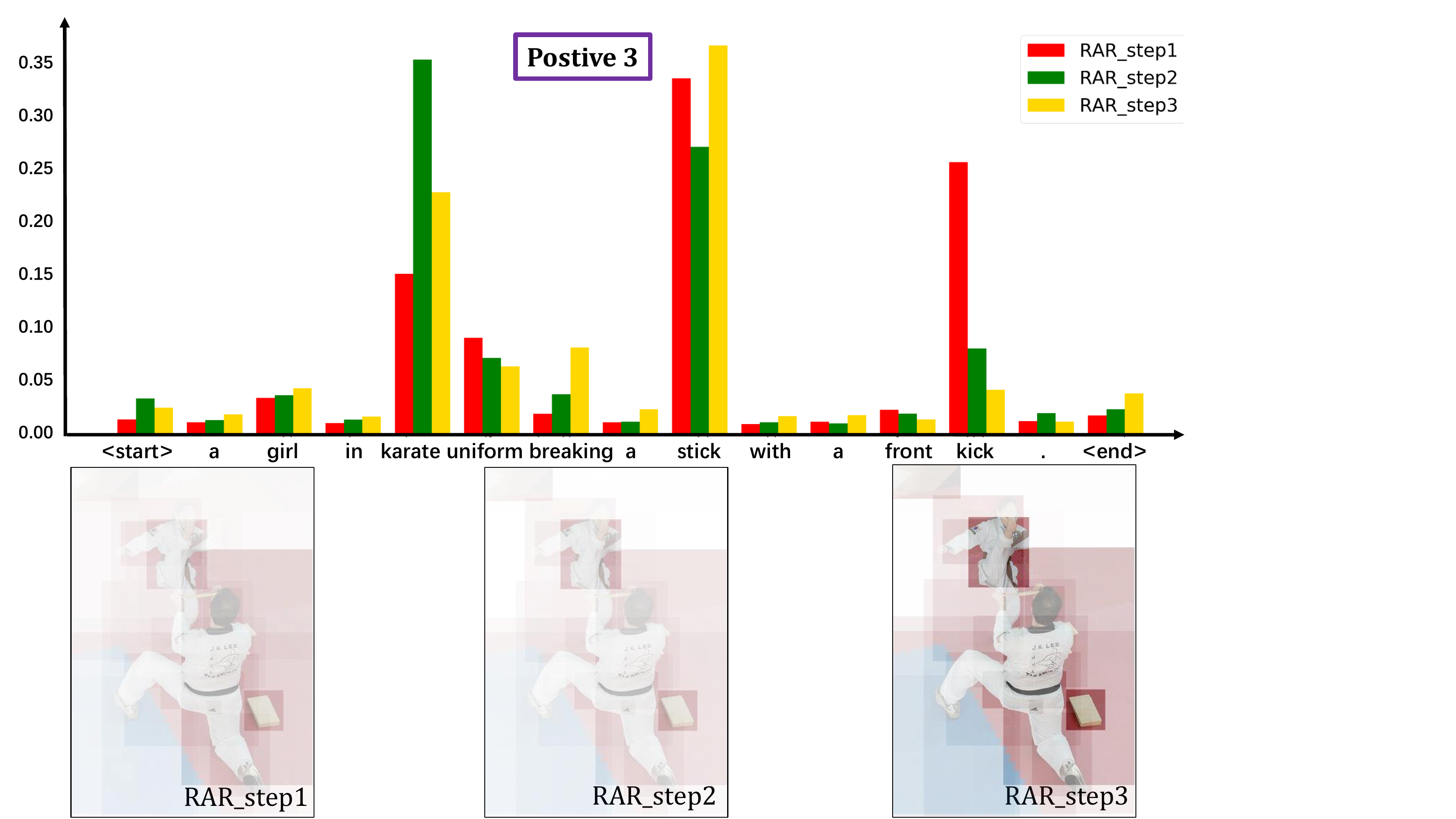}
		&\includegraphics[width=0.49\linewidth, height=0.32\linewidth, trim=0 0 170 0,clip]{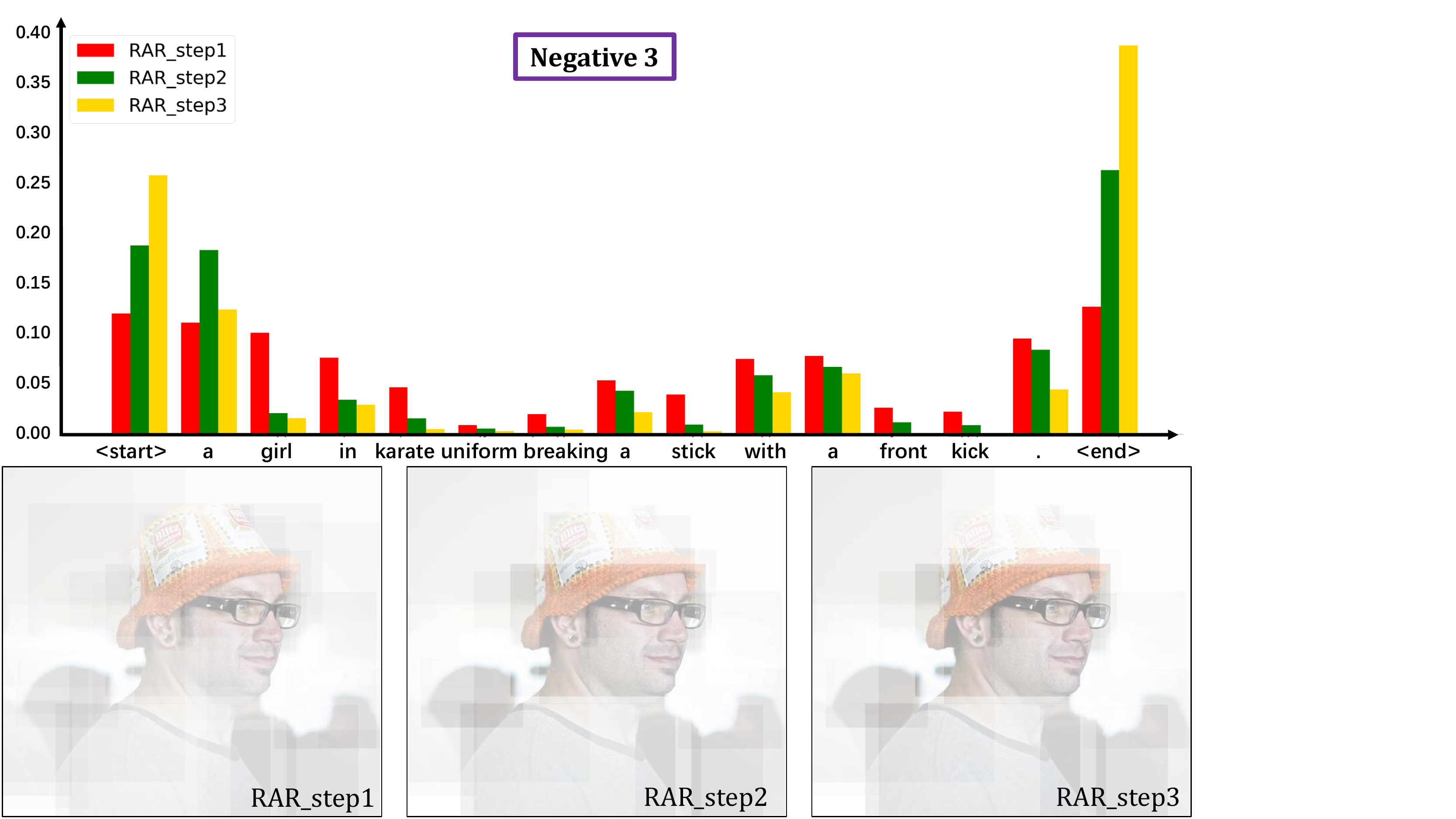} \\
	\end{tabular}
	\caption{Qualitative aggregation distribution by n-step RAR in the positive (right) and negative (left) pairs on Flickr30K. The histograms display the attention weights on word-based alignments with T2I attention while the images reflect the relative weights on region-based alignments with I2T attention.}
	\label{fig:RARsample}
\end{figure*}

\begin{figure*}[ht]
	\centering
	\begin{tabular}{@{}cc}
		\includegraphics[width=0.49\linewidth, height=0.20\linewidth,trim= 0 280 290 0,clip]{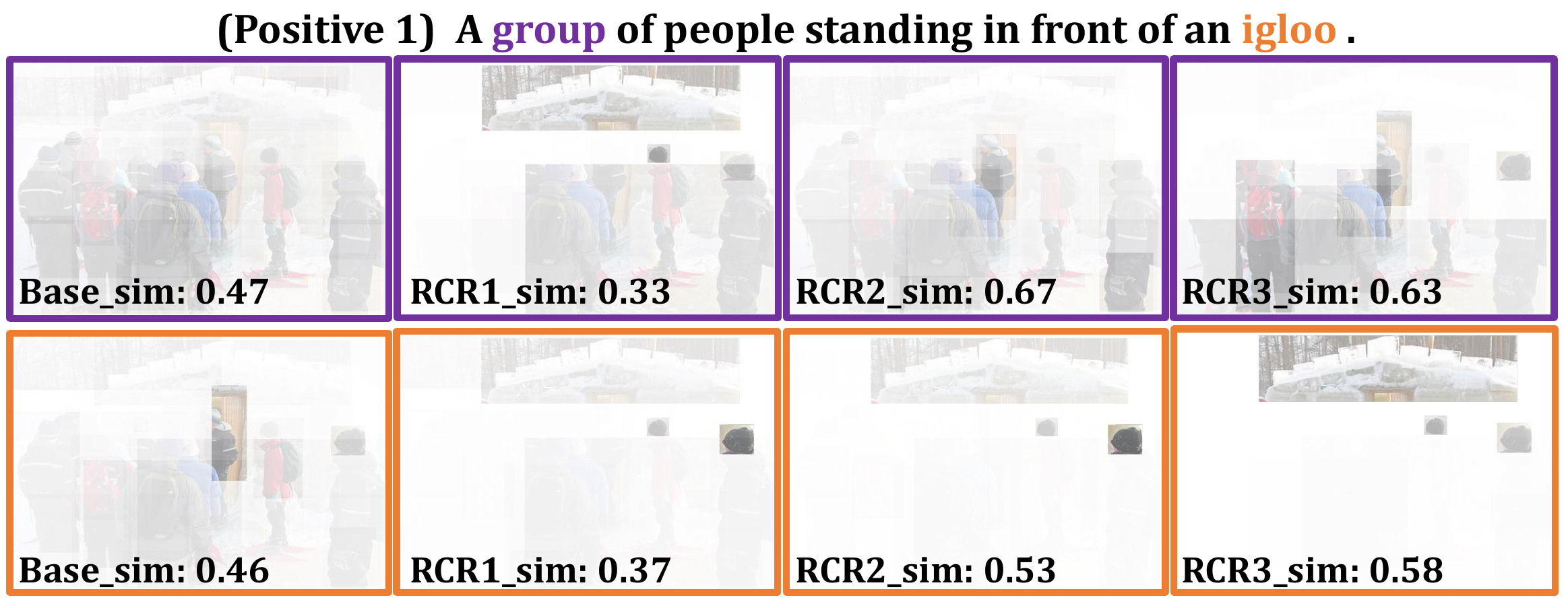}
		&\includegraphics[width=0.49\linewidth, height=0.20\linewidth,trim= 0 280 290 0,clip]{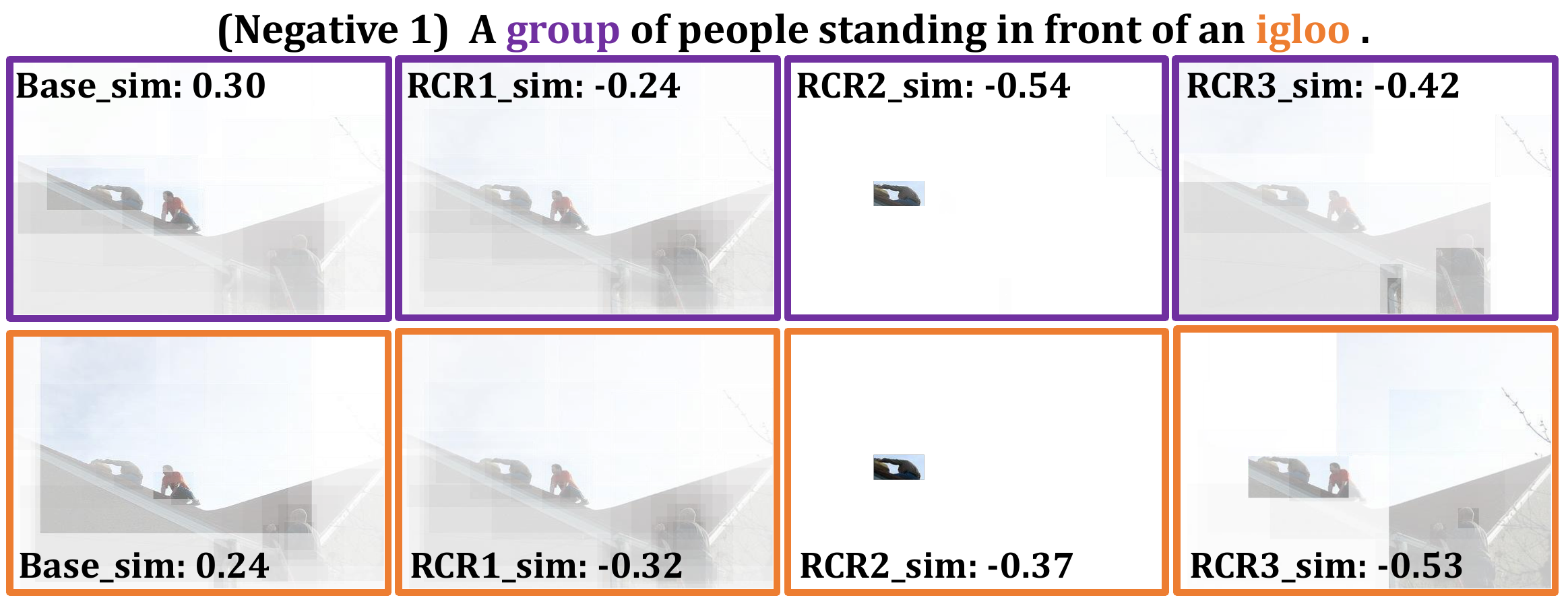}\\
		\includegraphics[width=0.49\linewidth, height=0.20\linewidth,trim= 0 280 290 0,clip]{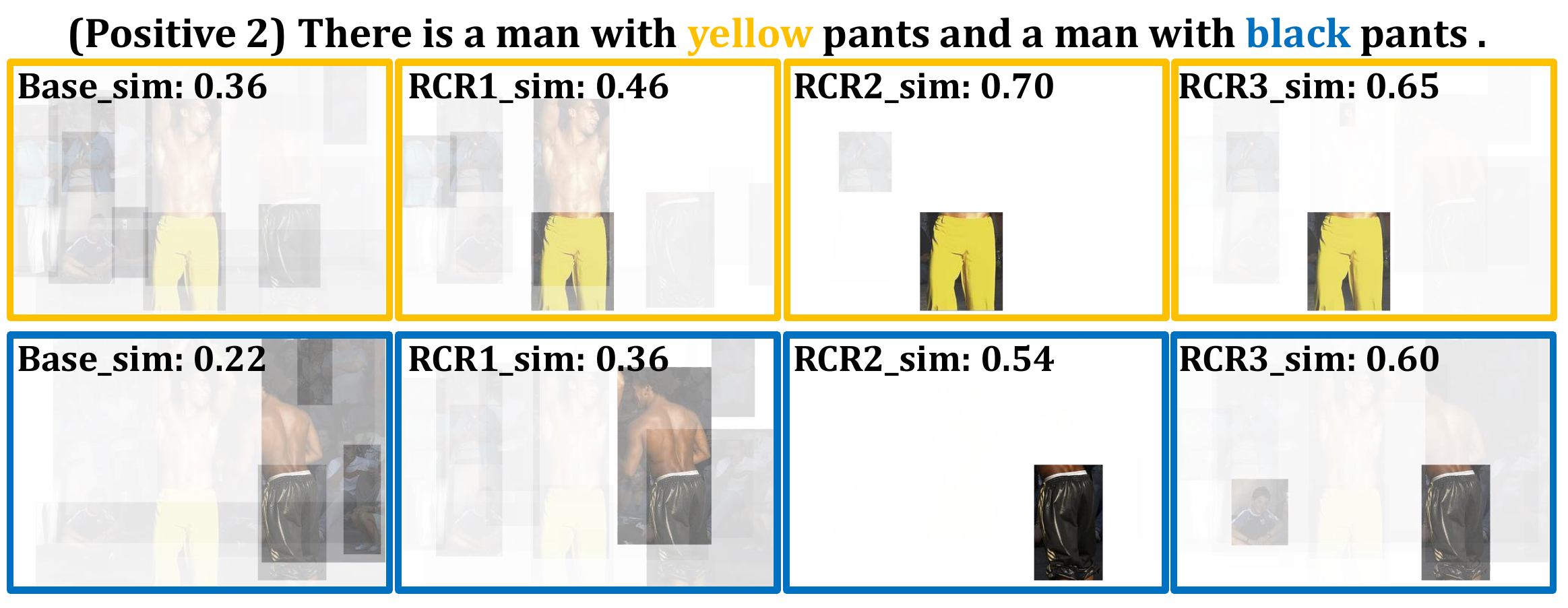}
		&\includegraphics[width=0.49\linewidth, height=0.20\linewidth,trim= 0 280 290 0,clip]{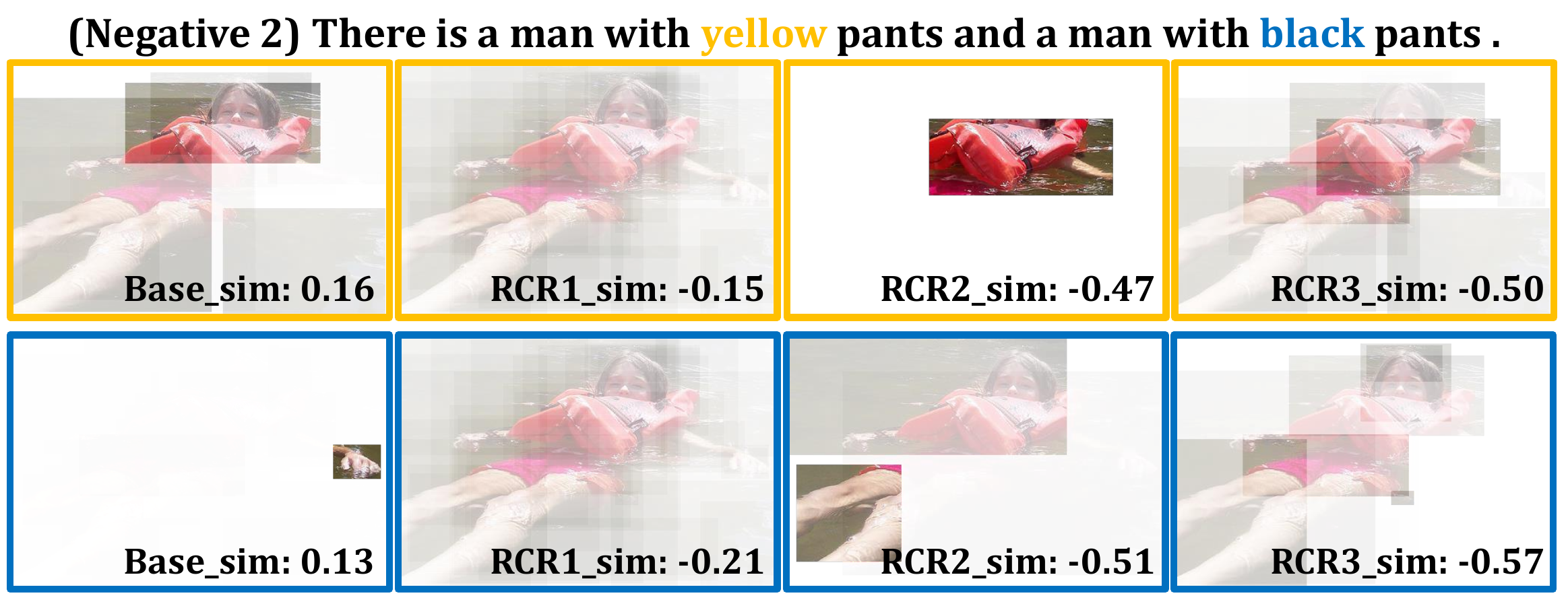}\\
		\includegraphics[width=0.49\linewidth, height=0.20\linewidth,trim= 0 280 290 0,clip]{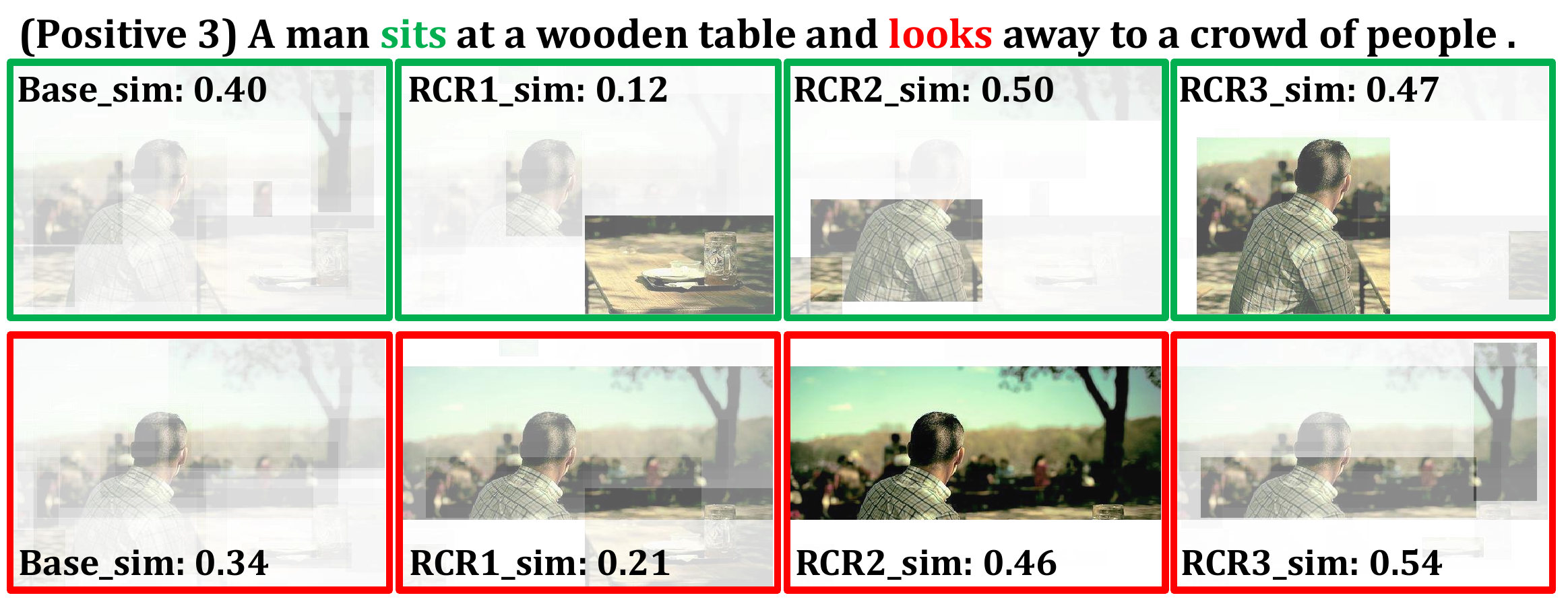}
		&\includegraphics[width=0.49\linewidth, height=0.20\linewidth,trim= 0 280 290 0,clip]{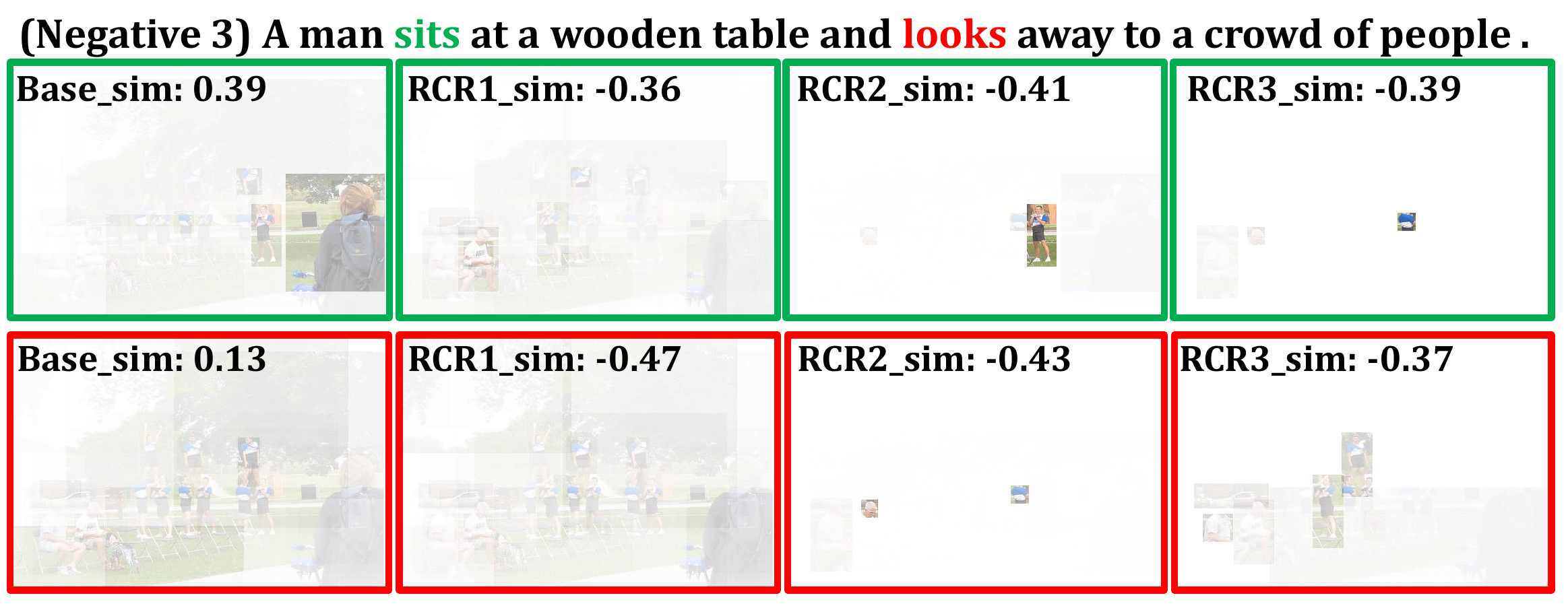}\\
	\end{tabular}
	\caption{Qualitative T2I attention distribution and word-based cosine similarities with diverse semantics by n-step RCR in the positive (right) and negative (left) pairs on Flickr30K. The image and $*\_$sim indicate the regions of interest and the corresponding cosine similarity according to the particular word.}
	\label{fig:RCRsample}
\end{figure*}

\textbf{Quantitative statistics of the regulators.}
Fig.~\ref{fig:RARattn} compares the attention weight distribution of different word-attended alignments with n-step RAR, while Fig.~\ref{fig:RCRdistance} displays the wasserstein distance of word-attended cosine similarities between positive and negative pairs with n-step RCR. Considering a very large range of vocabulary, we adopt NLTK toolkit and conduct statistical analyses with respect to the part of speech.
\textbf{1) Aggregation regulator.} For the matched image-text pairs, the RAR attempts to reduce the interference of less-meaningful alignments and highlight the important ones attended by nouns, adjectives, and verbs that contain rich semantic information. On the other hand, \texttt{"<start>"} and \texttt{"<end>"} encode the global textual contextual representations by Bi-GRU, and their corresponding alignments are emphasized gradually for the unmatched pairs which reflect the holistic differences 
across modalities. 
\textbf{2) Correspondence regulator.} From Fig.~\ref{fig:RCRdistance}, the mean distance learned by raw attention unit (T2I-SCAN~\cite{SCAN}) is nearly 0.35, and the one by the RCR is over 0.6 (step=2 best) with all parts of speech between positive and negative image-text pairs. We assume that attention weights are computed by one-step forward interactions with the fixed weight vector and softmax temperature, which obviously fail to measure feature channels and adapt itself to diverse words from different image-text pairs. Besides, even for the completely irrelevant image, the raw attention module still aligns the word with so-called "related regions" based on the cosine-like metrics and implicitly narrows the distances between the word and its related regions. In contrast, the RCR can make fine-grained adjustments with the prior alignments and refine the word-region correspondence progressively to produce larger gaps between matched and unmatched pairs.

\textbf{Qualitative aggregation distribution of n-step RAR.}
Fig.~\ref{fig:RARsample} illustrates the aggregation weights of word/region-attended alignments at the last step. We take Positive 1 as an example of positive pairs. The RAR with T2I attention can highlight the discriminative alignments (\textit{pierced ears, glasses, orange hat}) and abandon irrelevant ones (\textit{the, with, is}, etc.), while with I2T attention, it can also capture salient regions mentioned in the text (\textit{hat, ears, glasses}). Besides for negative pairs, the RAR with T2I attention tends to emphasize \textit{<start>}/\textit{<end>}-attended alignments which encode the overall image-text discrepancy. In terms of I2T attention, the aggregation distribution of image regions is relatively smooth to gain a more comprehensive prediction from the perspective of the image. As we can see, the RAR can selectively integrate important alignments and suppress less important ones.

\textbf{Qualitative cross-attention distribution of n-step RCR.}
Fig.~\ref{fig:RCRsample} exhibits the regions of interest and corresponding similarities with respect to the words with various semantics. Note that the final image-text score by the RCR is also computed by averaging all the word-attended cosine similarities. Hence, the similarity between a word and integrated regions can reflect their correspondence quantitatively (A higher score means a higher correlation, and vice versa). We can observe that the RCR can refine the word-region interactions step by step and gradually draw the distance between diverse words and their related regions for positive pairs. Compared with \textit{verbs} and \textit{adjectives}, \textit{nouns} are relatively easy to match for the Base model (T2I-SCAN~\cite{SCAN}). When the RCR is introduced, the nouns/verbs/adjectives-based correspondences become more accurate and fine-grained. More importantly, semantics-based similarities in negative pairs are pulled down significantly, indicating that plug-in RCR learns the difference and relationship from the previous alignments and reweighs the channel-wise and word-wise attention factors to associate words with ”completely irrelevant” regions in the latent space. By this means, the RCR can promote larger margins between positive and negative cross-modal pairs, and possess the greater capability to handle complex matching patterns.

\section{Conclusion and Future Works}
In this paper, we proposed two regulators termed as Recurrent Correspondence Regulator (RCR) and Recurrent Aggregation Regulator (RAR) to significantly facilitate the image-text matching process. Specifically, the RCR attempts to promote the cross-modal attention unit dynamically via learning more targeted attention factors, while the RAR aims to integrate the alignments progressively with plausible aggregation weights from holistic message feedback. 
The plug-and-play property enables them to seamlessly integrate into many existing approaches based on cross-modal interaction for achieving remarkable improvements, and more benefits can be obtained in a collaborative manner. 
Extensive experiments on MSCOCO and Flickr30K demonstrate the great superiority and broad applicability of our proposed approach.
Beyond the above observations, we also attempt to apply our regulators to another branch~\cite{VSE++,VSRN,SAEM,GPO} focusing on single-modality representations without cross-modality interactions. Interestingly, 2-step RCR and 3-step RAR can improve the R@1 of SAEM~\cite{SAEM} by 2.8/4.1\% and 3.7/2.5\% at two directions via gradually updating the last self-attention layer among regions and aggregating all instance features into a holistic feature respectively, reflecting the tremendous potential of our regulators. More efficient frameworks and application scenarios are one of our future research directions.
%%%------------------------------------------------------------------
\ifCLASSOPTIONcaptionsoff
  \newpage
\fi
\bibliographystyle{IEEEtran}
\bibliography{IEEEabrv,refs}
\end{document}